%% file: 43.tex
\documentclass[10pt,twocolumn,letterpaper]{article} 

\usepackage{etex}

\usepackage{avss}
\usepackage{times}
\usepackage{epsfig}
\usepackage{graphicx}
\usepackage{amsmath}
\usepackage{amssymb}

\usepackage{hyperref}

\usepackage{tikz}
\usetikzlibrary{shapes,arrows,positioning,calc,patterns,decorations.pathreplacing,angles,quotes}

\usepackage[export]{adjustbox}		% provides the key frame for \includegraphics

\usepackage[labelformat=simple]{subcaption}

\usepackage{bm}	% provides the bm command
\renewcommand{\vec}[1]{\bm{\mathbf{#1}}}

\avssfinalcopy % *** Uncomment this line for the final submission to remove line numbers

\def\avssPaperID{43} % *** Enter the AVSS Paper ID here

\ifavssfinal\fi
\begin{document}

%%%%%%%%% TITLE
\title{Stereo 3D Object Trajectory Reconstruction}

\author{Sebastian Bullinger, Christoph Bodensteiner, Michael Arens \\
Fraunhofer IOSB\\
Ettlingen, Germany\\
{\tt\small <firstname>.<lastname>@iosb.fraunhofer.de}
\and
Rainer Stiefelhagen\\
Karlsruhe Institute of Technology\\
Karslruhe, Germany\\
{\tt\small rainer.stiefelhagen@kit.edu}
}

\maketitle
% \thispagestyle{empty}

%The space after \eg, meaning ``for example'', should not be a sentence-ending space. So \eg is correct, {\em e.g.} is not.  The provided \verb'\eg' macro takes care of this.

%When citing a multi-author paper, you may save space by using ``et alia'', shortened to ``\etal'' (not ``{\em et.\ al.}'' as ``{\em et}'' is a complete word.)

% For this citation style, keep multiple citations in numerical (not chronological) order, so prefer \cite{Alpher03,Alpher02,Authors06} to \cite{Alpher02,Alpher03,Authors06}.

% Capitalize the first letter of nouns, pronouns, verbs, adjectives, and adverbs; do not capitalize articles, coordinate conjunctions, or prepositions (unless the title begins with such a word). 

\graphicspath{{images/}{images/dataset/}{images/overview/}{images/tracking/}{images/stereoConstraint/}{images/quantevaluation/}{images/qualevaluation/}{images/stuttgart00/}{images/stuttgart01/}{images/stuttgart02/}{images/kitti0013/}}

\input{abstract}

% Width and Height of grafics schon in overview 
\newlength{\includegraficsWidthOverview}
\setlength{\includegraficsWidthOverview}{1in}

\newlength{\includegraficsHeightOverview}
\setlength{\includegraficsHeightOverview}{0.5in}

\newlength{\subfigureWidthTwoColumns}
\setlength{\subfigureWidthTwoColumns}{2.4in}

\newlength{\subfigureWidthThreeColumns}
\setlength{\subfigureWidthThreeColumns}{1.6in}

\newlength{\subfigureWidthFourColumns}
\setlength{\subfigureWidthFourColumns}{1.65in}

\input{introduction}
\input{methods}

\input{experiments}

\input{conclusion}

{\small
\bibliographystyle{ieee}
\bibliography{egbib}
}

\end{document}

%% file: abstract.tex
\begin{abstract}
	We present a method to reconstruct the three-dimensional trajectory of a moving instance of a known object category using stereo video data. We track the two-dimensional shape of objects on pixel level exploiting instance-aware semantic segmentation techniques and optical flow cues. We apply Structure from Motion (SfM) techniques to object and background images to determine for each frame initial camera poses relative to object instances and background structures. We refine the initial SfM results by integrating stereo camera constraints exploiting factor graphs. We compute the object trajectory by combining object and background camera pose information. In contrast to stereo matching methods, our approach leverages temporal adjacent views for object point triangulation. As opposed to monocular trajectory reconstruction approaches, our method shows no degenerated cases. We evaluate our approach using publicly available video data of vehicles in urban scenes. 
\end{abstract}

%% file: introduction.tex
\section{Introduction}
\label{section:introduction}

\input{introduction_trajectory_reconstruction}

\input{introduction_related_work}

\subsection{Contribution}

The core contributions of this work are as follows.~(1) We present a new framework to reconstruct the three-dimensional trajectory of moving instances of known object categories in stereo video data leveraging state-of-the-art semantic segmentation and structure from motion approaches.~(2) We propose a novel approach to track the two-dimensional shape of objects on pixel level in stereo video data.~(3) We present a novel method to compute object motion trajectories consistent to image observations and background structures using state-of-the-art SfM techniques for data initialization and factor graphs for refinement exploiting stereo constraints.~(4) Opposed to stereo matching methods, our approach leverages views from different time steps for object point triangulation.~(5) We demonstrate the usefulness of our method by showing qualitative results of reconstructed object motion trajectories.

%\subsection{Paper Overview}
%TODO Add overview if 6 pages without references

%% file: introduction_trajectory_reconstruction.tex
\subsection{Trajectory Reconstruction}
\label{subsection:trajectory_reconstruction}

% augmented reality applications
The reconstruction of three-dimensional object motion trajectories is important for autonomous systems and surveillance applications. There are different platforms like drones or wearable systems where one wants to achieve this task with a minimal number of devices in order to reduce weight or lower production costs. We propose an approach to reconstruct three-dimensional object motion trajectories using two cameras as sensors. These results are essential for applications like environment perception and geo-registration of three-dimensional object trajectories. \\
3D stereo measurement precision deteriorates quickly with camera distance \cite{PinggeraECCV2014} due to limited camera baselines. We tackle this problem by combining temporal adjacent views using Structure from Motion techniques. Even small object rotations may result in big camera baseline differences. \\
In many scenes objects cover only a minority of pixels. This increases the difficulty of reconstructing object motion trajectories using image data. In such cases current state-of-the-art Structure from Motion (SfM) approaches \cite{Moulon2013,Schoenberger2016sfm} treat moving object observations most likely as outliers and reconstruct background structures instead. Previous works, e.g. \cite{Kundu2011, Lebeda2014}, detect moving objects by applying motion segmentation or keypoint tracking. Recent progress in instance-aware semantic segmentation \cite{Dai2016, HeGDG17} and optical flow \cite{Ilg2017, Hu2016} techniques allow for object tracking on pixel level \cite{Bullinger2017} and handle stationary objects naturally. We extend the approach in \cite{Bullinger2017} to track objects on pixel level in stereo video data. Stereo object tracking allows us to use \cite{Schoenberger2016sfm} and \cite{Moulon2013} for object and background reconstruction. We refine the reconstruction results by incorporating stereo constraints using GTSAM \cite{Daellert}. GTSAM provides functionality to model reconstruction problems with factor graphs. The incorporation of stereo constraints removes the scale ambiguity between object and background reconstruction and allows us to compute consistent object motion trajectories.

%% file: introduction_related_work.tex
\subsection{Related Work}
\label{subsection:related_work}

%\cite{CarloneICRA2015} show that the initialization using \cite{Martinec2007} results in high quality reconstruction results

Semantic segmentation or scene parsing is the task of providing semantic information at pixel-level. Shelhamer et al. \cite{Shelhamer2017} applied Fully Convolutional Networks for semantic segmentation, which are trained end-to-end. Recently, \cite{Dai2016,HeGDG17} proposed  instance-aware semantic segmentation approaches. The field of Structure from Motion (SfM) can be divided into iterative and global approaches. Iterative or sequential SfM methods \cite{Moulon2013,Schoenberger2016sfm,Sweeney2014} are more likely to find reasonable solutions than global SfM approaches \cite{Moulon2013, Sweeney2014}. However, the latter are less prone to drift. GTSAM \cite{Daellert} allows to model and to optimize SfM problems using factor graphs, but does not provide functionality to perform data association and initialization. \cite{CarloneICRA2015} analyze the importance of initialization techniques for Simultaneous Localization and Mapping (SLAM) using GTSAM. We perform data association and initialization using state-of-the-art SfM libraries \cite{Moulon2013,Schoenberger2016sfm}. Previous works \cite{Chhaya2016,Song2016} exploit specific camera poses to reconstruct object trajectories in monocular video data. These approaches are specifically defined for driving scenarios. \cite{EngelmannWACV2017} reconstruct vehicle shapes and trajectories in stereo video data using off-the-shelf ego-motion and stereo reconstruction algorithms. \cite{OsepICRA2017} combine object proposals, stereo, visual odometry and scene flow to compute three-dimensional vehicle tracks in traffic scenes. The object trajectory reconstructions in \cite{EngelmannWACV2017} and \cite{OsepICRA2017} are limited by the stereo camera baseline.

%% file: methods.tex
\section{Object Motion Trajectory Reconstruction}
\label{section:methods}

% 2
\input{methods_overview_figure}
\input{methods_object_motion_trajectory_computation_overview}

% 2.1

\input{methods_stereo_object_tracking_figure}
\input{methods_stereo_object_tracking}

% 2.2
\input{methods_object_motion_trajectory_computation}

% 2.3

\input{methods_object_motion_stereo_constraint_example_figure}

\input{experiments_trajectory_reconstruction_qual_figure}

%% file: methods_overview_figure.tex
\newcommand\firstcol{0}
\newcommand\firstrow{0}

\newcommand\secondcol{2.75}
\newcommand\secondrow{-3}

\newcommand\thirdcol{5.5}
\newcommand\thirdrow{-4.5}

\newcommand\includegraficsFirstRowDistanceRatio{0.5}
\newcommand\overviewVerticalShiftFirstRow{-0.25}
\newcommand\overviewVerticalShiftSecondToThirdRow{0.25}
\newcommand\overviewVerticalShiftThirdRow{-0.25}

\begin{figure}[!tb]
	\centering
	
	% available sizes are:
	% \tiny, \scriptsize, \footnotesize, \small, \normalsize, \large, \Large, \LARGE, \huge, \Huge 
	\newcommand{\figureTextSize}{\footnotesize}
	
	%http://tex.stackexchange.com/questions/147143/whats-the-difference-between-path-and-draw-in-tikz
	% To do an invisible path we use \path and if you want to put some ink on it you use \draw

	% Define block styles
	% Further possible options: scale=0.8
	\tikzstyle{blockRounded} = [rectangle, draw, text width=6em, text centered, rounded corners, font=\figureTextSize]
	\tikzstyle{blockCorner} = [rectangle, draw, text width=6em, text centered, font=\figureTextSize]
	\tikzstyle{line} = [draw, -latex', font=\figureTextSize, text centered, text width=18mm]
	
	\tikzstyle{imageStyle} = [% postaction={fill, red}, % For debugging purposes
							text width = \includegraficsWidthOverview,
							text height = \includegraficsHeightOverview]
	\tikzstyle{imageDescriptionStyle} = [% postaction={fill, red}, % For debugging purposes,
										fill=white, 
										opacity=0.5, 
										text opacity=1, 
										font=\figureTextSize,
										text width = \includegraficsWidthOverview]
	\tikzstyle{plainDescriptionStyle} = [
		rectangle,
		draw,
		fill=white, 
		opacity=1, 
		text opacity=1, 
		font=\figureTextSize,
		text width = \includegraficsWidthOverview]
	
	% other options:
	% text badly centered
	% inner sep=1pt
	
	  \begin{tikzpicture}[auto]

	  % coordinates are (x,y) = (width, height)
	  
	  \coordinate (A) at (-1.5, - 0.8);
	  \coordinate (B) at ($(A)+(4.25, 0)$);
	  \coordinate (C) at ($(B)+(0, -2.25)$);
	  \coordinate (D) at ($(C)+(1.35, 0)$);
	  \coordinate (E) at ($(D)+(0, -2.25)$);
	  \coordinate (F) at ($(A)+(0, -4.5)$);
	  
	   \node [text=blue, align=left] (perObject) at ($(A)+(0.75,-0.5)$) {Per \\ Object};
	  
	  % options: dash dot, dashed, dotted
	  \draw[thick,dashed,blue] (A) -- (B) -- (C) -- (D) -- (E) -- (F) -- cycle;

		\node (inputFrames) at (\firstcol,\firstrow) {};
	  	\node [blockRounded] (semanticSegmentation)  at (\secondcol,\firstrow)  {Semantic Segmentation and Object Tracking};
		\node [blockRounded, text width=5em] (objectSfM)  at (\firstcol,\secondrow) {SfM + Stereo Refinement};
	  	\node [blockRounded, text width=5em] (backgroundSfM)  at (\thirdcol,\secondrow)  {SfM + Stereo Refinement};

	    % connecting nodes with paths and use line (defined previously) as style
	  \begin{scope}[every path/.style=line]
	  % paths for enrollemnt rows
	  \draw (inputFrames) -- (semanticSegmentation);
	  \node[imageStyle] (framesToTracking) at (inputFrames) { 
	  	\includegraphics[ frame, width = \includegraficsWidthOverview]{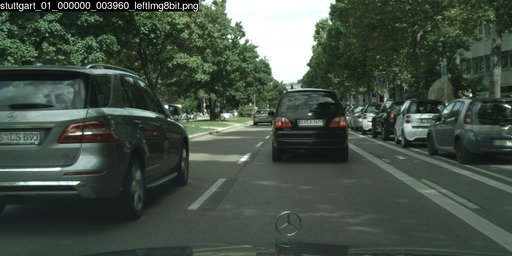}};
	  \node[imageDescriptionStyle] at (framesToTracking.center) {Input Frames (Left + Right)};

	  \draw (semanticSegmentation) -- 
	  (objectSfM);
	  \node[imageStyle] (sSegmentationToObject) at ($(semanticSegmentation)!\includegraficsFirstRowDistanceRatio!(objectSfM)+(0,\overviewVerticalShiftFirstRow)$) { 
	  	\includegraphics[ frame, width = \includegraficsWidthOverview]{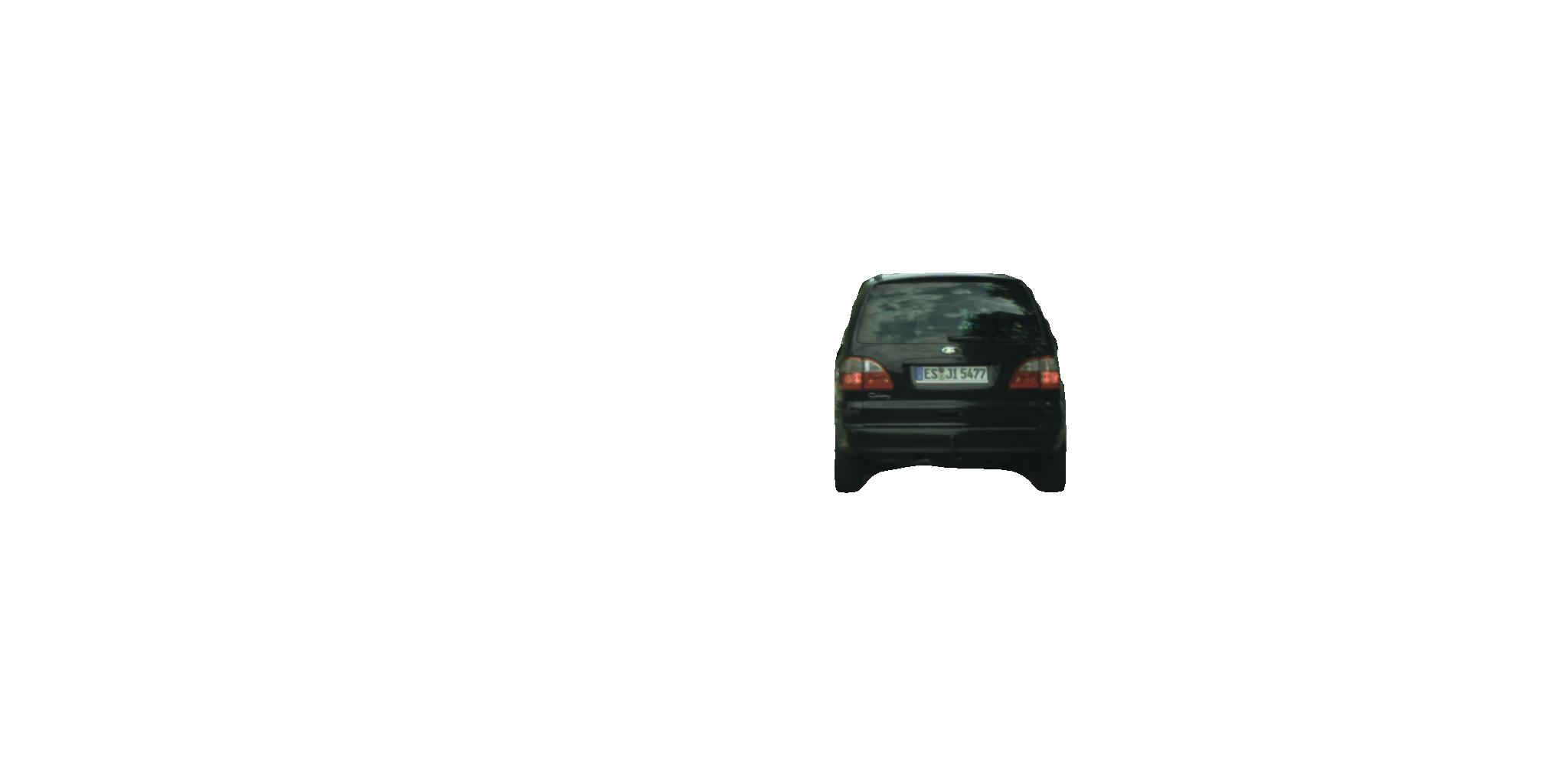}};
	  \node[imageDescriptionStyle] at (sSegmentationToObject.center) {Object \\ Segmentations (Left + Right)};
	  
	  \draw (semanticSegmentation) -- (backgroundSfM);
	  \node[imageStyle] (sSegmentationToBackgroundImage) at  ($(semanticSegmentation)!\includegraficsFirstRowDistanceRatio!(backgroundSfM)+(0,\overviewVerticalShiftFirstRow)$){ 
	  	\includegraphics[ frame, width = \includegraficsWidthOverview]{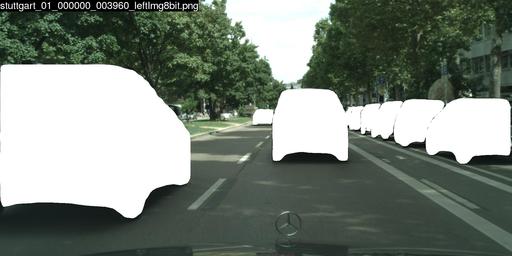}};
	  \node[imageDescriptionStyle] (sSegmentationToBackgroundDescription) at (sSegmentationToBackgroundImage.center) {Background Segmentations \\ (Left + Right)};
	  
	  \node[imageStyle] (oSfmToFamImage) at ($(\firstcol,\thirdrow)$){ 
		\includegraphics[frame, width = \includegraficsWidthOverview]{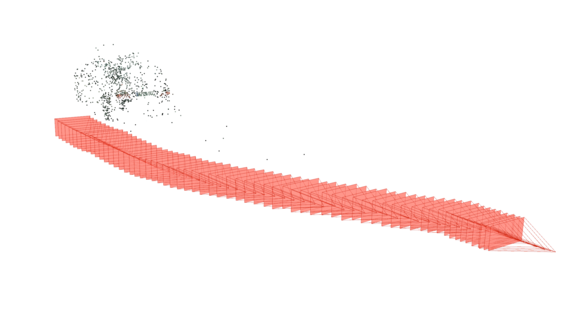}};
	   \node[imageDescriptionStyle] at (oSfmToFamImage.center) {Object \\ SfM Result};

	  \node[imageStyle] (bSfmToFamImage) at (\thirdcol,\thirdrow) { 
	  	\includegraphics[frame, width = \includegraficsWidthOverview]{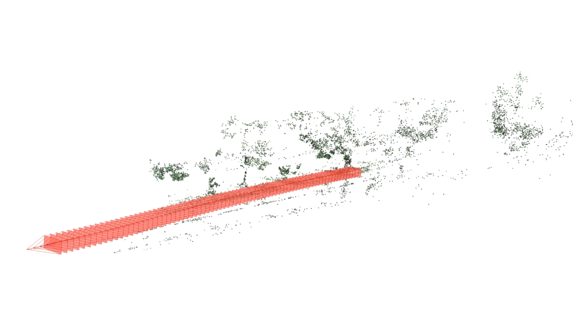}};
	  \node[imageDescriptionStyle] at (bSfmToFamImage.center) {Background SfM Result};

	  \node [blockRounded, text width=4.5em] (trajectoryComputation) at (\secondcol,\thirdrow) {Trajectory Computation};
	  
	  \draw (backgroundSfM) -- (bSfmToFamImage) -- (trajectoryComputation);
	  \draw (objectSfM) -- (oSfmToFamImage) -- (trajectoryComputation);
	  
	  \end{scope}

	  % Add description to images
  		% node[left, pos=0.4, fill=white, opacity=0.5, text opacity=1] {Background SfM Result}
	  
	  \end{tikzpicture}
	
	%\newlength{\flowChartVerticalLength}
	%\setlength{\flowChartVerticalLength}{0.5cm}
	
	%\newlength{\flowChartHorizontalLength}
	%\setlength{\flowChartHorizontalLength}{-0.1cm}
	
	% http://www.texample.net/tikz/examples/simple-flow-chart/
	%\begin{tikzpicture}[node distance = \flowChartVerticalLength]
	%==== For simple graphs ... ====
	%\node [blockCorner] (inputFrames) {Input Frames};
	%\node [blockRounded , below = of inputFrames] %(semanticSegmentation) {Semantic Segmentation};
	%\node [blockCorner, below left = \flowChartVerticalLength and %\flowChartHorizontalLength of semanticSegmentation] %(objectImages) {Object Images};
	% ================================
	
	%\end{tikzpicture}
	
	\caption{Overview of the Trajectory Reconstruction Pipeline. Boxes with corners denote computation results and boxes with rounded corners denote computation steps.}
	\label{methods:overview}
\end{figure}
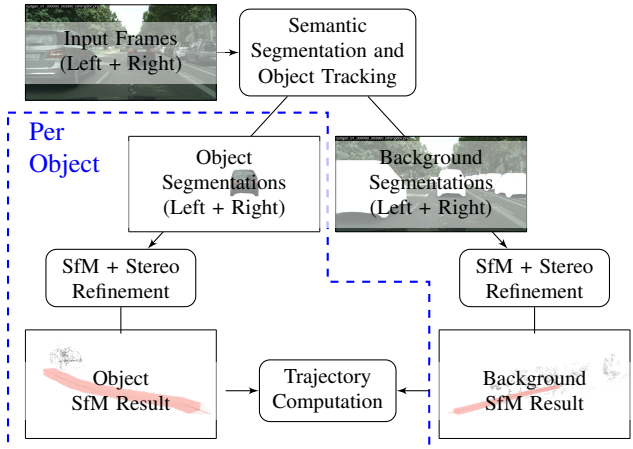

%% file: methods_object_motion_trajectory_computation_overview.tex
%\subsection{Object Motion Trajectory Computation}
%\label{subsection:trajectory_computation}

The pipeline of our approach is shown in Fig.~\ref{methods:overview}. The input is an ordered stereo image sequence. We track two-dimensional object shapes on pixel level across video sequences exploiting instance-aware semantic segmentation \cite{Li2016} to identify object shapes and optical flow \cite{Ilg2017} to associate extracted object shapes in corresponding stereo images and subsequent frames. Without loss of generality, we describe motion trajectory reconstructions of single objects. We apply SfM \cite{Moulon2013,Schoenberger2016sfm} to object and background images as shown in Fig.~\ref{methods:overview}. Object images denote pictures containing only color information of single object instances. Similarly, background images show only environment structures. We combine information of object and background SfM reconstructions to determine consistent object motion trajectories. We use GTSAM \cite{Daellert} to refine object and background reconstructions and resolve the scale ambiguity using stereo camera baseline constraints.\\
The point triangulation of stereo matching or stereo correspondence \cite{Scharstein2002} methods are limited by the baseline of corresponding the stereo camera \cite{PinggeraECCV2014}. In contrast, SfM allows to triangulate 3D points by exploiting information of subsequent frames. Since already small object rotations may result in big camera baseline changes, our method is not necessarily limited by the stereo camera baseline. In contrast to stereo correspondence methods, the proposed approach builds object models reflecting the information of each frame. To build an object model with stereo matching techniques requires additional steps to fuse the 3D points of subsequent frames. The presented method does not require a calibration of the stereo camera.

%% file: methods_stereo_object_tracking_figure.tex
\begin{figure*}[!htb]
	% available sizes are:
	% \tiny, \scriptsize, \footnotesize, \small, \normalsize, \large, \Large, \LARGE, \huge, \Huge 
	\newcommand{\figureTextSize}{\footnotesize}
	
	\centering
	\begin{tikzpicture}[auto]
	
		% https://tex.stackexchange.com/questions/15028/calculating-right-angle-triangle-side-inside-latex
		% \pgfmathsetmacro and \pgfmathsetlengthmacro can also be defined outside of tikzppicture
		% \pgfmathsetmacro and \pgfmathsetlengthmacro supports floating point numbers
		% \pgfmathsetmacro esults in a numeric output which is void of any unit of measure
		% \pgfmathsetlengthmacro preserves the dimension component
		
		\pgfmathsetlengthmacro{\includegraficsWidthOF}{1.1in}
		
		\pgfmathsetmacro{\colDistance}{3.5}
		\pgfmathsetmacro{\rowDistance}{-2.25}
		
		\coordinate (fullRow) at (0, \rowDistance);
		\coordinate (fullCol) at (\colDistance, 0);
		
		\coordinate (halfRow) at (0,0.5 * \rowDistance);
		\coordinate (halfCol) at (0.5 * \colDistance, 0);
		
		\coordinate (diagLineOffsetLD) at (-0.2,-0.2);
		\coordinate (diagLineOffsetRD) at (-0.2,0.2);
		
		\pgfmathsetmacro{\firstcol}{0}
		\pgfmathsetmacro{\firstrow}{0}
		
		\pgfmathsetmacro{\secondcol}{\colDistance}
		\pgfmathsetmacro{\secondrow}{\rowDistance}
		
		\pgfmathsetmacro{\thirdcol}{2 * \colDistance}
		\pgfmathsetmacro{\thirdrow}{2 * \rowDistance}
	
		\pgfmathsetmacro{\fourthcol}{3 * \colDistance}
		\pgfmathsetmacro{\fourthrow}{3 * \rowDistance}
		
		\pgfmathsetmacro{\fifthcol}{4 * \colDistance}
		\pgfmathsetmacro{\fifthrow}{4 * \rowDistance}
		
		\pgfmathsetmacro{\sixthcol}{5 * \colDistance}
		\pgfmathsetmacro{\sixthrow}{5 * \rowDistance}
		
		\pgfmathsetmacro{\bracerow}{5.3 * \rowDistance}
		\pgfmathsetmacro{\braceWidthSingle}{0.85*\colDistance}
		\pgfmathsetmacro{\braceWidthDouble}{1.9*\colDistance}
		\pgfmathsetmacro{\firstbracecolumn}{-0.5 * \colDistance}
		\pgfmathsetmacro{\secondbracecolumn}{1.5 * \colDistance}
		\pgfmathsetmacro{\thirdbracecolumn}{3.5 * \colDistance}

		\draw[thick,dashed,blue] 
			($ (\firstcol, \firstrow) -(halfRow) +(halfCol) $) -- 
			++($2.5*(fullRow)$) -- 
			++($0.95*(fullCol)$) --
			++($3.3*(fullRow)$);
			
		\draw[thick,dashed,blue] 
			($ (\thirdcol, \firstrow) -(halfRow) +(halfCol) $) -- 
			++($2.5*(fullRow)$) -- 
			++($0.95*(fullCol)$) --
			++($3.3*(fullRow)$);

		% coordinates are (x,y) = (width, height)
		\coordinate (descr_offset) at (0,1);
	  
	  	% === Input Images Left ===
 		\node (imageLeftI) at (\firstcol,\firstrow) { 
	  	\includegraphics[ frame, width = \includegraficsWidthOF]{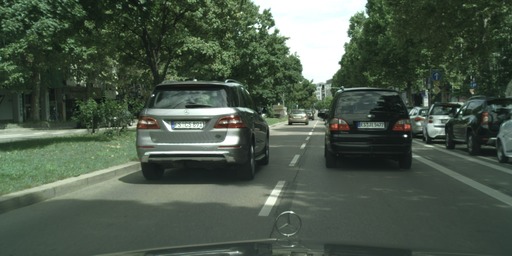}};
		
 		\node (imageLeftI1) at (\thirdcol,\firstrow) { 
		\includegraphics[ frame, width = \includegraficsWidthOF]{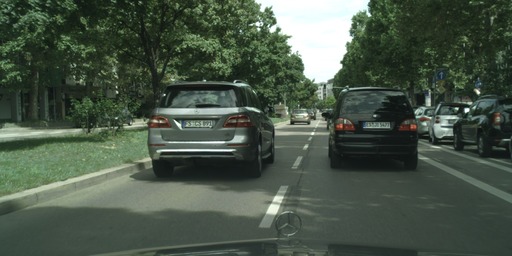}};

 		\node (imageLeftI2) at (\fifthcol,\firstrow) { 
		\includegraphics[ frame, width = \includegraficsWidthOF]{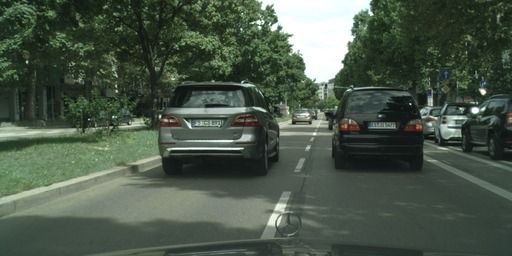}};

		% === Tracker State images Left ===
 		\node (trackerLeftI) at (\firstcol,\thirdrow) { 
			\includegraphics[ frame, width = \includegraficsWidthOF]{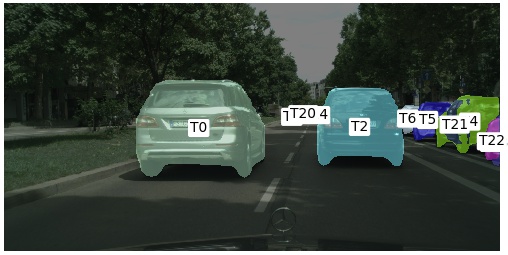}};
		
 		\node (trackerLeftI1) at (\thirdcol,\thirdrow) { 
			\includegraphics[ frame, width = \includegraficsWidthOF]{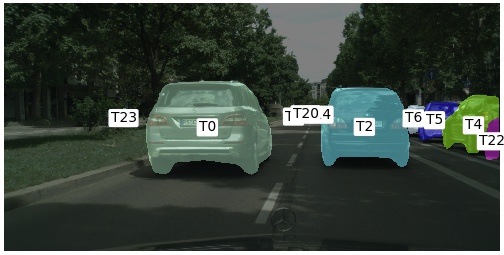}};
		
 		\node (trackerLeftI2) at (\fifthcol,\thirdrow) { 
			\includegraphics[ frame, width = \includegraficsWidthOF]{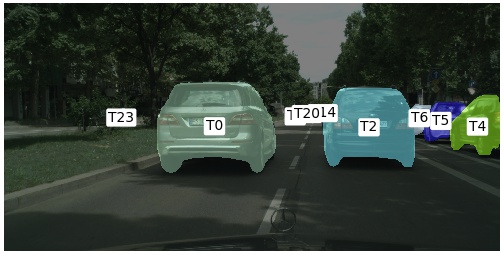}};

		% === Left Detections Images ===
		\node (detectionLeftI) at (\firstcol,\secondrow){ 
			\includegraphics[frame, width = \includegraficsWidthOF]{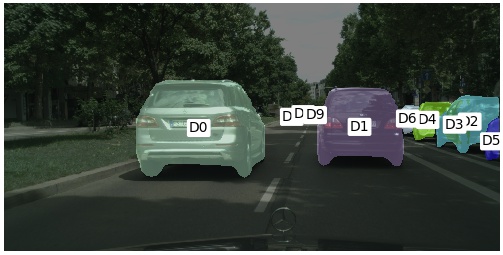}};
		
		\node (detectionLeftI1) at (\thirdcol,\secondrow){ 
			\includegraphics[frame, width = \includegraficsWidthOF]{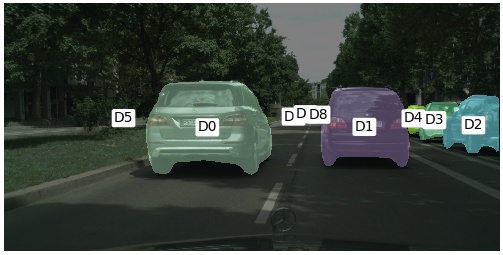}};
		
		\node (detectionLeftI2) at (\fifthcol,\secondrow){ 
			\includegraphics[frame, width = \includegraficsWidthOF]{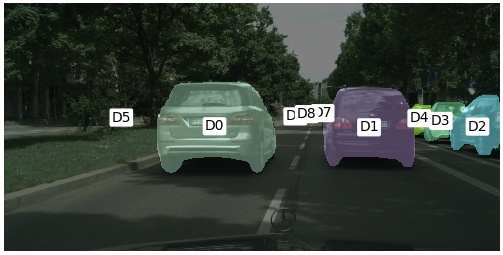}};
		
		% === Left Prediction Images ===
		\node (predictionLeftI) at ($(\secondcol, \secondrow) + (halfRow)$){ 
			\includegraphics[frame, width = \includegraficsWidthOF]{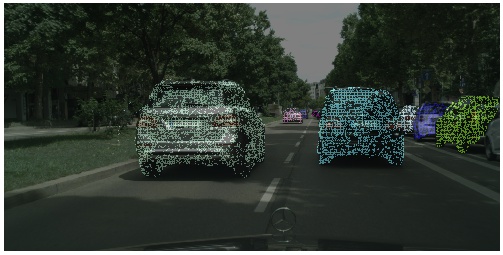}};
		
		\node (predictionLeftI1) at ($(\fourthcol, \secondrow) + (halfRow)$){ 
			\includegraphics[frame, width = \includegraficsWidthOF]{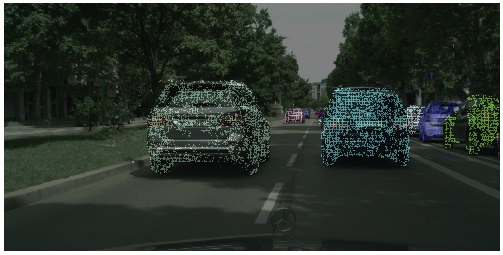}};

		% === LN Optical Flow Images

		\node (OF_LN) at ($(\secondcol, \firstrow)  + (halfRow)$){ 
			\includegraphics[frame, width = \includegraficsWidthOF]{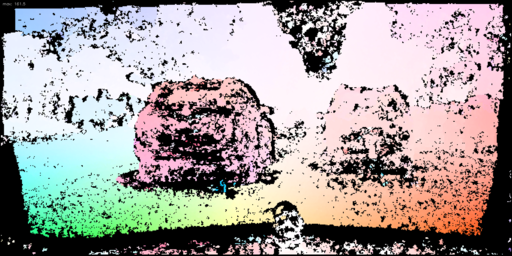}};
		\node (OF_LN1) at ($(\fourthcol, \firstrow) + (halfRow)$){ 
		\includegraphics[frame, width = \includegraficsWidthOF]{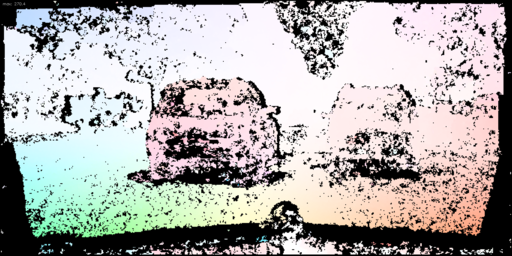}};

		% === LR Prediction Images
		\node (predictionRightI) at ($(\firstcol, \fourthrow)$){ 
			\includegraphics[frame, width = \includegraficsWidthOF]{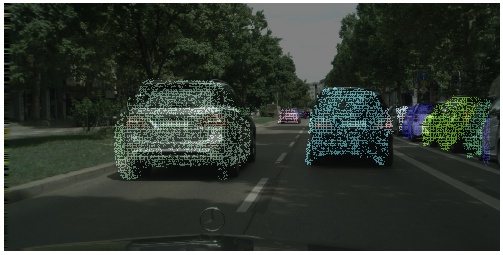}};
		\node (predictionRightI1) at ($(\thirdcol, \fourthrow)$){ 
			\includegraphics[frame, width = \includegraficsWidthOF]{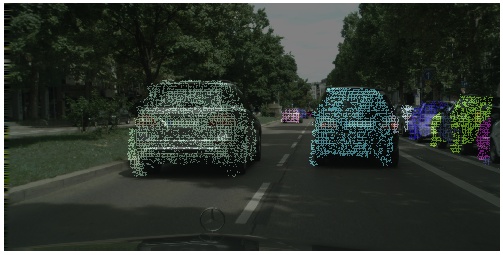}};
		\node (predictionRightI2) at ($(\fifthcol, \fourthrow)$){ 
			\includegraphics[frame, width = \includegraficsWidthOF]{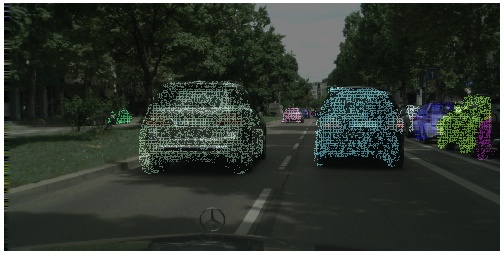}};

		% === LR Optical Flow
		\node (OF_LR) at  (\firstcol,\fifthrow) { 
			\includegraphics[ frame, width = \includegraficsWidthOF]{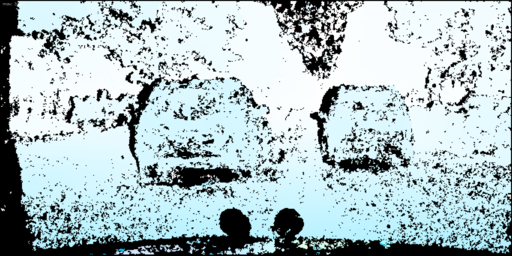}};
		\node (OF_LR1) at  (\thirdcol,\fifthrow) { 
			\includegraphics[ frame, width = \includegraficsWidthOF]{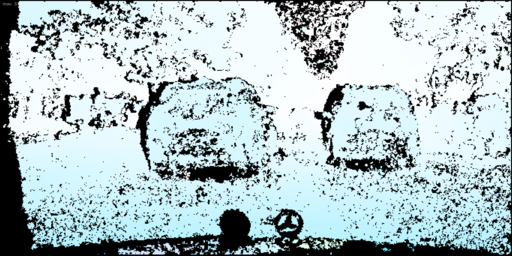}};
		\node (OF_LR2) at  (\fifthcol,\fifthrow) { 
			\includegraphics[ frame, width = \includegraficsWidthOF]{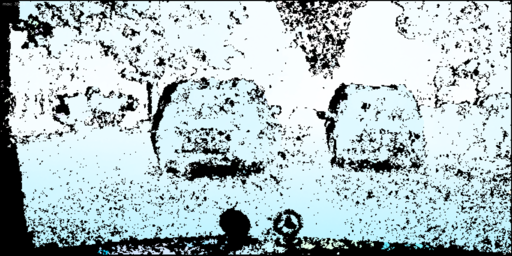}};
		
		% === Input Images Right 
		\node (imageRightI) at  (\firstcol,\sixthrow) { 
			\includegraphics[ frame, width = \includegraficsWidthOF]{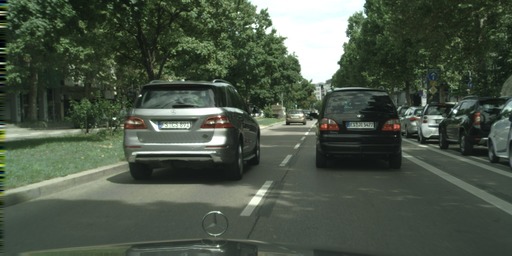}};
		
		\node (imageRightI1) at  (\thirdcol,\sixthrow) { 
			\includegraphics[ frame, width = \includegraficsWidthOF]{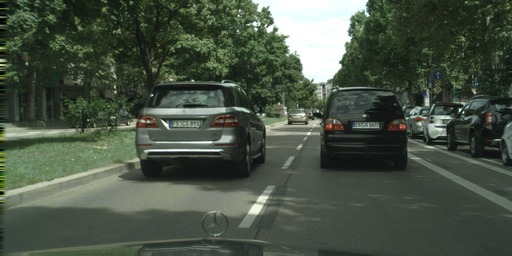}};
		
		\node (imageRightI2) at  (\fifthcol,\sixthrow) { 
			\includegraphics[ frame, width = \includegraficsWidthOF]{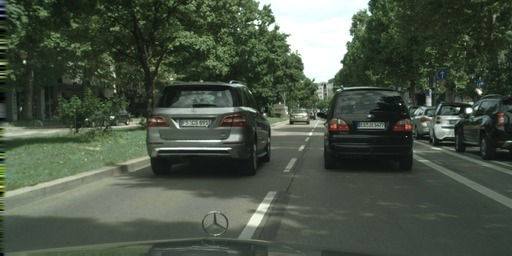}};

		% === Right Detections Images ===
		\node (detectionRightI) at (\secondcol,\fifthrow){ 
			\includegraphics[frame, width = \includegraficsWidthOF]{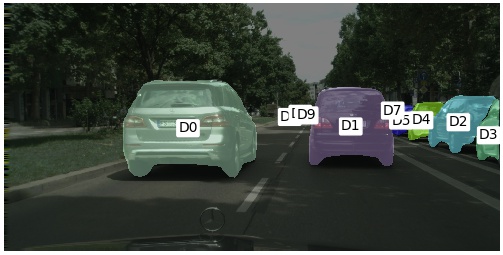}};
		
		\node (detectionRightI1) at (\fourthcol,\fifthrow){ 
			\includegraphics[frame, width = \includegraficsWidthOF]{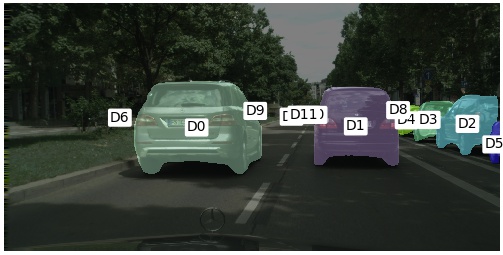}};

		% === Tracker State images Right ===
		\node (trackerRightI) at (\secondcol,\fourthrow) { 
			\includegraphics[ frame, width = \includegraficsWidthOF]{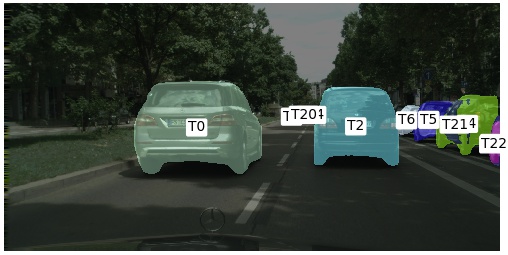}};
		
		\node (trackerRightI1) at (\fourthcol,\fourthrow) { 
			\includegraphics[ frame, width = \includegraficsWidthOF]{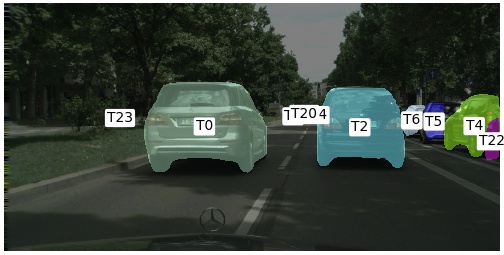}};

	  	% === Optical Flow LN Lines (In) ===
	  	\tikzset{arrowStyle/.style={->, line width=0.5mm}}
		\draw[arrowStyle] (imageLeftI) -- (OF_LN);
		\draw[arrowStyle] (imageLeftI1) -- (OF_LN);
		\draw[arrowStyle] (imageLeftI1) -- (OF_LN1);
		\draw[arrowStyle] (imageLeftI2) -- (OF_LN1);

		% === Segmentation Lines Left (In) ===
		\draw[arrowStyle] (imageLeftI) -- (detectionLeftI);
		\draw[arrowStyle] (imageLeftI1) -- (detectionLeftI1);
		\draw[arrowStyle] (imageLeftI2) -- (detectionLeftI2);
		
		% === Segmentation Lines Right (In) ===
		\draw[arrowStyle] (imageRightI) -- (detectionRightI);
		\draw[arrowStyle] (imageRightI1) -- (detectionRightI1);
		
		% === Tracker Lines Left (In) ===
		\draw[arrowStyle] (detectionLeftI) -- (trackerLeftI);
		\draw[arrowStyle] (detectionLeftI1) -- (trackerLeftI1);
		\draw[arrowStyle] (detectionLeftI2) -- (trackerLeftI2);
		
		\draw[arrowStyle] (predictionLeftI) -- (trackerLeftI1);
		\draw[arrowStyle] (predictionLeftI1) -- (trackerLeftI2);
		
	    % === Tracker Lines Right (In) ===
		\draw[arrowStyle] (detectionRightI) -- (trackerRightI);
		\draw[arrowStyle] (detectionRightI1) -- (trackerRightI1);

		\draw[arrowStyle] (predictionRightI) -- (trackerRightI);
		\draw[arrowStyle] (predictionRightI1) -- (trackerRightI1);

		% === Prediction Lines Left (In) ===
		\draw[arrowStyle] (OF_LN) -- (predictionLeftI);
		\draw[arrowStyle] (OF_LN1) -- (predictionLeftI1);
		
		\draw[arrowStyle] (trackerLeftI) -- (predictionLeftI);
		\draw[arrowStyle] (trackerLeftI1) -- (predictionLeftI1);
		
	  	% === Optical Flow LR Lines (In) ===
	  	
		\draw[arrowStyle] (imageLeftI.south west) -- ($(imageLeftI.south west) +(diagLineOffsetLD)$) -- ($(OF_LR.north west) + (diagLineOffsetRD)$) -- (OF_LR.north west);
		\draw[arrowStyle] (imageRightI) -- (OF_LR);

		\draw[arrowStyle] (imageLeftI1.south west) -- ($(imageLeftI1.south west) +(diagLineOffsetLD)$) -- ($(OF_LR1.north west) + (diagLineOffsetRD)$) -- (OF_LR1.north west);
		\draw[arrowStyle] (imageRightI1) -- (OF_LR1);
		
		\draw[arrowStyle] (imageLeftI2.south west) -- ($(imageLeftI2.south west) +(diagLineOffsetLD)$) -- ($(OF_LR2.north west) + (diagLineOffsetRD)$) -- (OF_LR2.north west);
		\draw[arrowStyle] (imageRightI2) -- (OF_LR2);
		
		% === Prediction Lines Right (In) ===
		\draw[arrowStyle] (OF_LR) -- (predictionRightI);
		\draw[arrowStyle] (OF_LR1) -- (predictionRightI1);
		\draw[arrowStyle] (OF_LR2) -- (predictionRightI2);
		
		\draw[arrowStyle] (trackerLeftI) -- (predictionRightI);
		\draw[arrowStyle] (trackerLeftI1) -- (predictionRightI1);
		\draw[arrowStyle] (trackerLeftI2) -- (predictionRightI2);

		\tikzset{descrStyle/.style={}}
		% === Input Image Descriptions Left
		\node [descrStyle] at ($(imageLeftI)+(descr_offset)$) {$I_{i,l}$};
		\node [descrStyle] at ($(imageLeftI1)+(descr_offset) $) {$I_{i+1,l}$};
		\node [descrStyle] at ($(imageLeftI2)+(descr_offset) $) {$I_{i+2,l}$};
		
		% === Input Image Descriptions Right
		\node [descrStyle, right] at ($(imageRightI)+(descr_offset) $) {$I_{i,r}$};
		\node [descrStyle, right] at ($(imageRightI1)+(descr_offset) $) {$I_{i+1,r}$};
		\node [descrStyle, right] at ($(imageRightI2)+(descr_offset) $) {$I_{i+2,r}$};

		% === Optical Flow Descriptions Left
		\node [descrStyle] at ($(OF_LN)+(descr_offset) $) {$OF_{i,ln}$};
		\node [descrStyle] at ($(OF_LN1)+(descr_offset) $) {$OF_{i+1,ln}$};
		
		% === Optical Flow Descriptions Right
		\node [descrStyle, right] at ($(OF_LR)+(descr_offset) $) {$OF_{i,lr}$};
		\node [descrStyle, right] at ($(OF_LR1)+(descr_offset) $) {$OF_{i+1,lr}$};
		\node [descrStyle, right] at ($(OF_LR2)+(descr_offset) $) {$OF_{i+2,lr}$};

		% === Detection Descriptions Left
		\node [descrStyle, right] at ($(detectionLeftI)+(descr_offset)$) {$D_{i,l}$};
		\node [descrStyle, right] at ($(detectionLeftI1)+(descr_offset)$) {$D_{i+1,l}$};
		\node [descrStyle, right] at ($(detectionLeftI2)+(descr_offset)$) {$D_{i+2,l}$};
		
		% === Detection Descriptions right
		\node [descrStyle, right] at ($(detectionRightI)+(descr_offset)$) {$D_{i,r}$};
		\node [descrStyle, right] at ($(detectionRightI1)+(descr_offset)$) {$D_{i+1,r}$};
		
		% === Prediction Descriptions Left
		\node [descrStyle, right] at ($(predictionLeftI)+(descr_offset)$) {$P_{i,ln}$};
		\node [descrStyle, right] at ($(predictionLeftI1)+(descr_offset)$) {$P_{i+1,ln}$};
		
		% === Prediction Descriptions Right
		\node [descrStyle, right] at ($(predictionRightI)+(descr_offset)$) {$P_{i,lr}$};
		\node [descrStyle, right] at ($(predictionRightI1)+(descr_offset)$) {$P_{i+1,lr}$};
		\node [descrStyle, right] at ($(predictionRightI2)+(descr_offset)$) {$P_{i+2,lr}$};
		
		% === Tracklet Descriptions Left
		\node [descrStyle, right] at ($(trackerLeftI)+(descr_offset)$) {$T_{i,l}$};
		\node [descrStyle, right] at ($(trackerLeftI1)+(descr_offset)$) {$T_{i+1,l}$};
		\node [descrStyle, right] at ($(trackerLeftI2)+(descr_offset)$) {$T_{i+2,l}$};
		
		% === Tracklet Descriptions Right
		\node [descrStyle, right] at ($(trackerRightI)+(descr_offset)$) {$T_{i,r}$};
		\node [descrStyle, right] at ($(trackerRightI1)+(descr_offset)$) {$T_{i+1,r}$};

		\draw[decoration={brace,mirror,raise=5pt},decorate]
		($(\firstbracecolumn,\bracerow)$) -- node[below=6pt] {$t_i$} ($(\firstbracecolumn+\braceWidthDouble,\bracerow)$);

		\draw[decoration={brace,mirror,raise=5pt},decorate]
		($(\secondbracecolumn,\bracerow)$) -- node[below=6pt] {$t_{i+1}$} ($(\secondbracecolumn+\braceWidthDouble,\bracerow)$);

		\draw[decoration={brace,mirror,raise=5pt},decorate]
		($(\thirdbracecolumn,\bracerow)$) -- node[below=6pt] {$t_{i+2}$} ($(\thirdbracecolumn+\braceWidthSingle,\bracerow)$);

	  \end{tikzpicture}

	\caption{Stereo Object Tracking Scheme. The variables have the following meaning. $I$: image, $OF$: optical flow, $D$: detection, $P$: Prediction, $T$: Tracker State, $i$: image index, $l$: left, $r$: right, $ln$: left-next, $lr$: left-right. Arrows show the relation of computation steps. A computation step depends on the results connected with incoming arrows. The optical flow color coding used is defined in \cite{Baker2007}. The figure is best viewed in color.}
	 % The blue lines separate results using information of different time steps. 
	\label{methods:stereo_tracking_schemes}
\end{figure*}
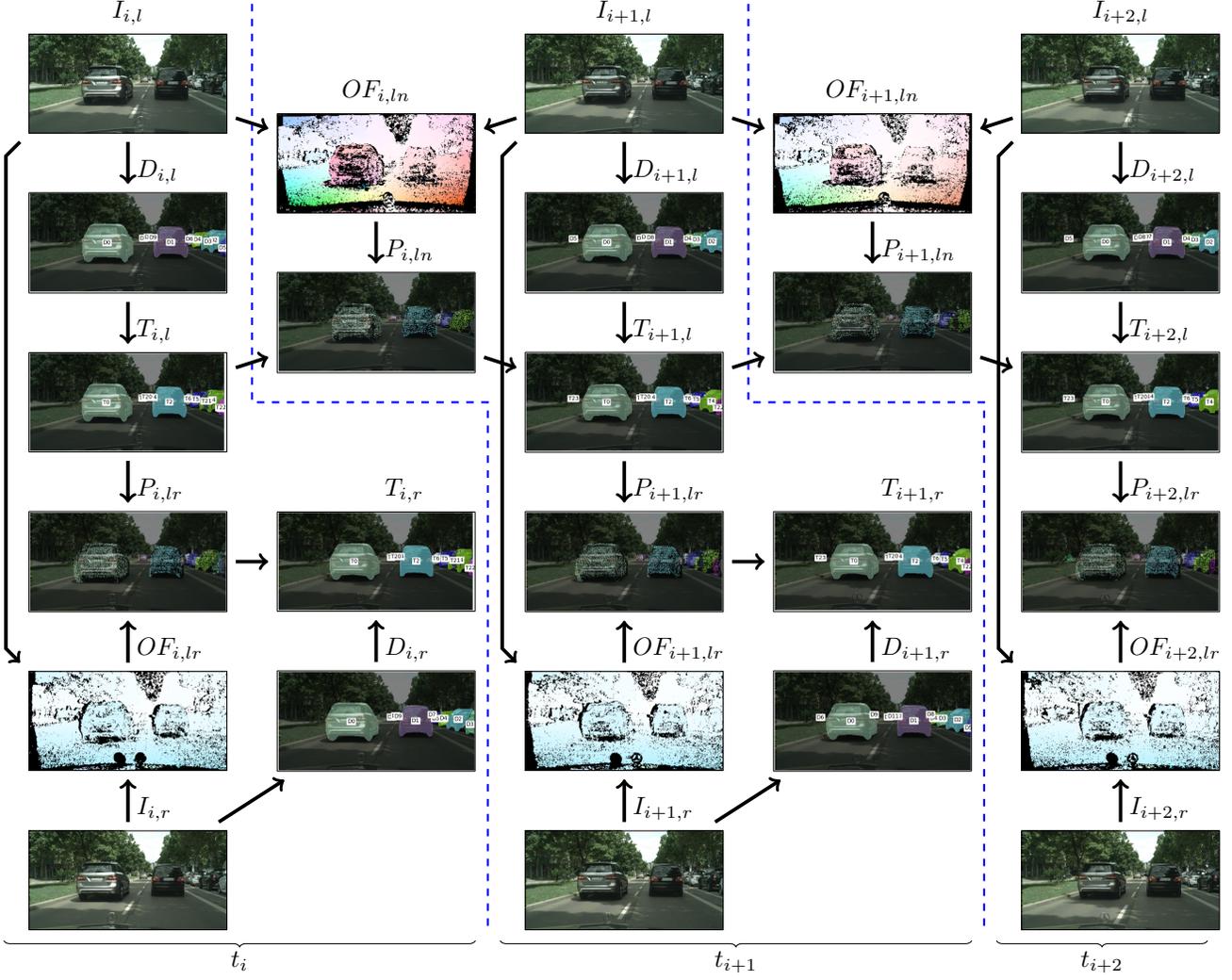

%% file: methods_stereo_object_tracking.tex
\subsection{Stereo Online Multiple Object Tracking}

%% https://en.wikipedia.org/wiki/Assignment_problem
%%consists of finding a maximum weight matching (or minimum weight perfect matching) in a weighted bipartite graph. 

%% http://www.cse.ust.hk/~golin/COMP572/Notes/Matching.pdf
%
%Bipartite Graph Matching in $\mathcal{O}(n)^3$, where $n$ is the number of nodes. Bipartite Matching is also refered to as 2-dimensional matching. \\
%In a stereo camera setting one has to find object proposals in subsequent left and right stereo images. 
%% https://en.wikipedia.org/wiki/Multipartite_graph
%% TODO mention multipartite graphs
%This turns the correspondence problem into a 4-dimensional matching problem. However, the optimazation of a 4-dimensional matching is known to be NP-hard problem.
%% https://en.wikipedia.org/wiki/3-dimensional_matching#CITEREFCrescenziKannHalld%C3%B3rssonKarpinski2000
%% Since the decision problem described above is NP-complete, this optimization problem is NP-hard, and hence it seems that there is no polynomial-time algorithm for finding a maximum 3-dimensional matching
%Therefore, we apply the following greedy algorithm scheme to determine object proposal correspondences.
%
%
%%https://www.topcoder.com/community/data-science/data-science-tutorials/assignment-problem-and-hungarian-algorithm/

% The recent progress in instance-aware semantic segmentation \cite{HeGDG17} and optical flow \cite{Hu2016} allows to compute both 
The proposed Stereo Multiple Object Tracking (MOT) approach extends the monocular tracking algorithm presented in \cite{Bullinger2017} and is depicted in Fig.~\ref{methods:stereo_tracking_schemes}. \cite{Bullinger2017} allows to track the two-dimensional shape of objects of known categories across video sequences on pixel level. We use optical flow matches to associate instance-aware semantic segmentations between subsequent frames to maintain the tracker state. In contrast to motion model based tracking methods, this approach allows to naturally associate objects between left and right images of stereo cameras. \cite{Bullinger2017} uses the Kuhn-Munkres algorithm \cite{Kuhn1955} to solve the assignment problem, i.e. to determine object associations of objects between image pairs. The assignment problem consists of finding a maximum weight matching in a weighted two-dimensional (or bipartite) graph. This problem translates in the stereo MOT case to a four-dimensional matching problem, because the object instances in the left image $I_{i,l}$ and the right image $I_{i,r}$ at time $i$ as well as the object instances in the left image $I_{i+1,l}$ and the right image $I_{i+1,r}$ at time $i+1$ must be associated. Let $OF_{i,lr}$ and $OF_{i,ln}$ denote the optical flow between image $I_{i,l}$ and $I_{i,r}$ as well as $I_{i,l}$ and $I_{i+1,l}$. We do not solve the associations of $I_{i,l}$, $I_{i+1,l}$, $I_{i,r}$ and $I_{i+1,r}$ simultaneously, since (a) the four-dimensional matching problem is NP-complete and (b) the simultaneous determination of two subsequent stereo image pairs requires the computation of three optical flow fields in addition to $OF_{i,lr}$ and $OF_{i,ln}$. Instead, we track object instances in the left images $I_{i,l}$ and $I_{i+1,l}$ using the object affinity matrix presented \cite{Bullinger2017} as input for the Kuhn-Munkres algorithm. Concretely, the affinity matrix is defined according to equation \eqref{eq:affinity_matrix}
\begin{equation}
\label{eq:affinity_matrix}
\vec{A}_{t} =
\begin{bmatrix}
O_{1,1} 	& \cdots & O_{1,v} 		& \cdots & O_{1,n_v} \\
\vdots 		& \ddots & \vdots 		& \ddots & \vdots \\
O_{u,1} 	& \cdots & O_{u,v} 		& \cdots & O_{u,n_v} \\
\vdots 		& \ddots & \vdots 		& \ddots & \vdots \\
O_{n_u,1} 	& \cdots & O_{n_u,v}	& \cdots & O_{n_u,n_v}
\end{bmatrix}
.
\end{equation}
Here, $O_{u,v}$ denotes the overlap of prediction $u$ in $P_{i,ln}$ and detection $v$ in $D_{i+1,l}$ (see Fig.~\ref{methods:stereo_tracking_schemes}). Let $n_u$ denote the number of predictions in $P_{i,ln}$ and $n_v$ denote the number of detections in $D_{i+1,l}$. This affinity measure reflects locality and visual similarity. The tracker state $T_{i+1,l}$ contains only tracks of object instances in images corresponding to the left camera. We use the optical flow between left and right images $OF_{i+1,lr}$ to associate the tracker state of left images $T_{i+1,l}$ with objects visible in the corresponding right image. The association between predictions $P_{i+1,lr}$ and detections $D_{i+1,r}$ in the right images are also solved using the affinity matrix of \cite{Bullinger2017} as input for \cite{Kuhn1955}. In this case $O_{u,v}$ denotes the overlap of prediction $u$ in $P_{i+1,lr}$ and detection $v$ in $D_{i+1,r}$. $n_u$ denotes the number of predictions in $P_{i+1,lr}$ and $n_v$ denotes the number of detections in $D_{i+1,r}$.

%% file: methods_object_motion_trajectory_computation.tex
\subsection{Object Motion Trajectory Computation}
\label{subsection:trajectory_computation}
We follow the pipeline outlined in Fig.~\ref{methods:overview} and apply SfM simultaneously to object and background images. We denote corresponding reconstruction results with $sfm^{(o)}$ and $sfm^{(b)}$. Each object image has a corresponding background image, i.e. the background image extracted from the same input frame. We consider only object-background-image-pairs, which are part of $sfm^{(o)}$ and $sfm^{(b)}$. Reconstructed cameras without corresponding object or background camera are removed from the reconstruction. \\
Let $\vec{o}_{j}^{(o)}$ denote the 3D points contained in $sfm^{(o)}$. The superscript $o$ in $\vec{o}_{j}^{(o)}$ describes the corresponding coordinate frame. The variable $j$ denotes the index of the points in the object point cloud. We combine information of object-background-image-pairs to define object motion trajectories parameterized by a single parameter. The object reconstruction $sfm^{(o)}$ contains object point positions $\vec{o}_{j}^{(o)}$ as well as corresponding camera centers $\vec{c}_{i}^{(o)}$ and rotations $\vec{R}_{i}^{(o)}$. We convert the object points $\vec{o}_{j}^{(o)}$ defined in the coordinate frame system (CFS) of the object reconstruction to points in the camera CFS $\vec{o}_{j}^{(i)}$ of camera $i$ using $\vec{o}_{j}^{(i)} = \vec{R}_{i}^{(o)} \cdot (\vec{o}_{j}^{(o)} - \vec{c}_{i}^{(o)})$. We use the camera center $\vec{c}_{i}^{(b)}$ and the corresponding rotation $\vec{R}_{i}^{(b)}$ contained in the background reconstruction $sfm^{(b)}$ to transform object points in camera coordinates to the background CFS using equation \eqref{eq:camera_to_background_coordinates}.
\begin{equation}
\label{eq:camera_to_background_coordinates}
\vec{o}_{j,i}^{(b)} = \vec{c}_{i}^{(b)} + {\vec{R}_{i}^{(b)}}^T \cdot \vec{o}_{j}^{(i)}
\end{equation}
The naive combination of object and background reconstruction results in inconsistent object motion trajectories due to the scale ambiguity of SfM \cite{Hartley2004}. We adjust the scale between object and background reconstruction using the baseline of the stereo cameras in object and background reconstructions as reference. Reconstructions of dynamic objects using state-of-the-art SfM tools contain occasionally badly registered cameras and incorrectly triangulated object points (see Fig.~\ref{methods:registration_error_example}). Reasons for these are small object sizes, changing illumination and reflecting surfaces. Incorrectly estimated camera baselines hamper the correct estimation of the scale ratio between object and background reconstruction. \\
We leverage factor graphs \cite{Daellert} to model stereo camera constraints and to refine the previously computed SfM reconstructions. For each triangulated point we search for corresponding stereo feature observations, i.e. pairs of feature observations which appear in the left and the right image of the same time step. Stereo image rectification preprocessing allows us to assume that the stereo feature observation positions should show (almost) the same $y$ coordinate. Since the feature observations are computed for each image independently we use only stereo feature observations with an $y$ pixel difference smaller than three pixels. We average the $y$ coordinate to define the final stereo constraint. The resulting reconstructions show consistent camera stereo baselines. Note that GTSAM \cite{Daellert} does not provide functionality to perform data association and initialization. Fig.~\ref{methods:registration_error_example} shows a comparison of initial and refined reconstructions. \\
We can recover the full object motion trajectory computing equation \eqref{eq:camera_to_background_coordinates} for each object-background-image-pair. We use $\vec{o}_{j,i}^{(b)}$ of all cameras and object points as object motion trajectory representation.

%% file: methods_object_motion_stereo_constraint_example_figure.tex
\newlength{\subfigureWidthTwoColumnsInSingleColumn}
\setlength{\subfigureWidthTwoColumnsInSingleColumn}{1.6in}

\begin{figure}[!tb]
	% use the subfloat package (not subfigure or subfig)
	\begin{subfigure}[t]{0.2\textwidth}
		\includegraphics[width=\subfigureWidthTwoColumnsInSingleColumn,frame]{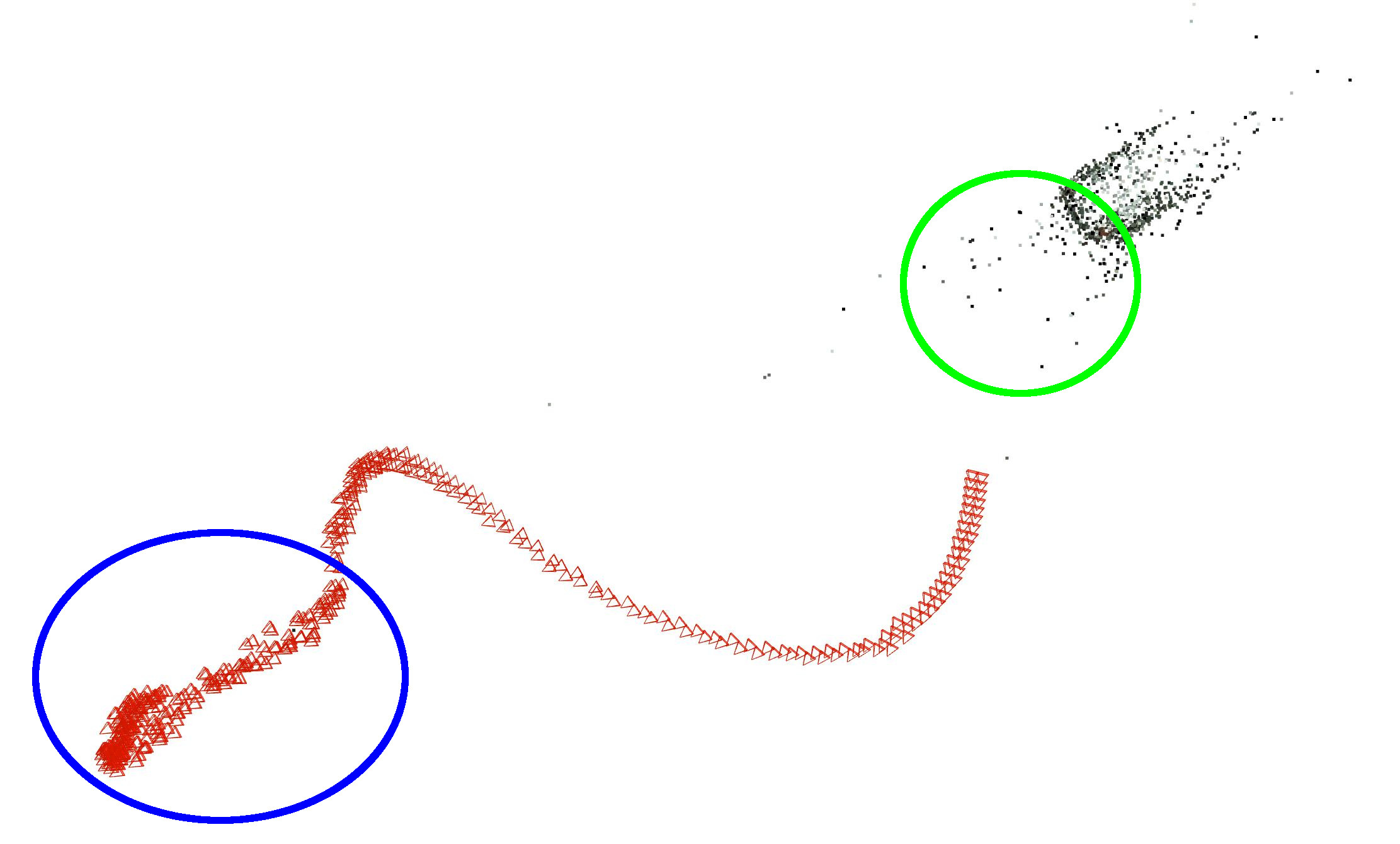}
		\hfill
		\includegraphics[width=\subfigureWidthTwoColumnsInSingleColumn,frame]{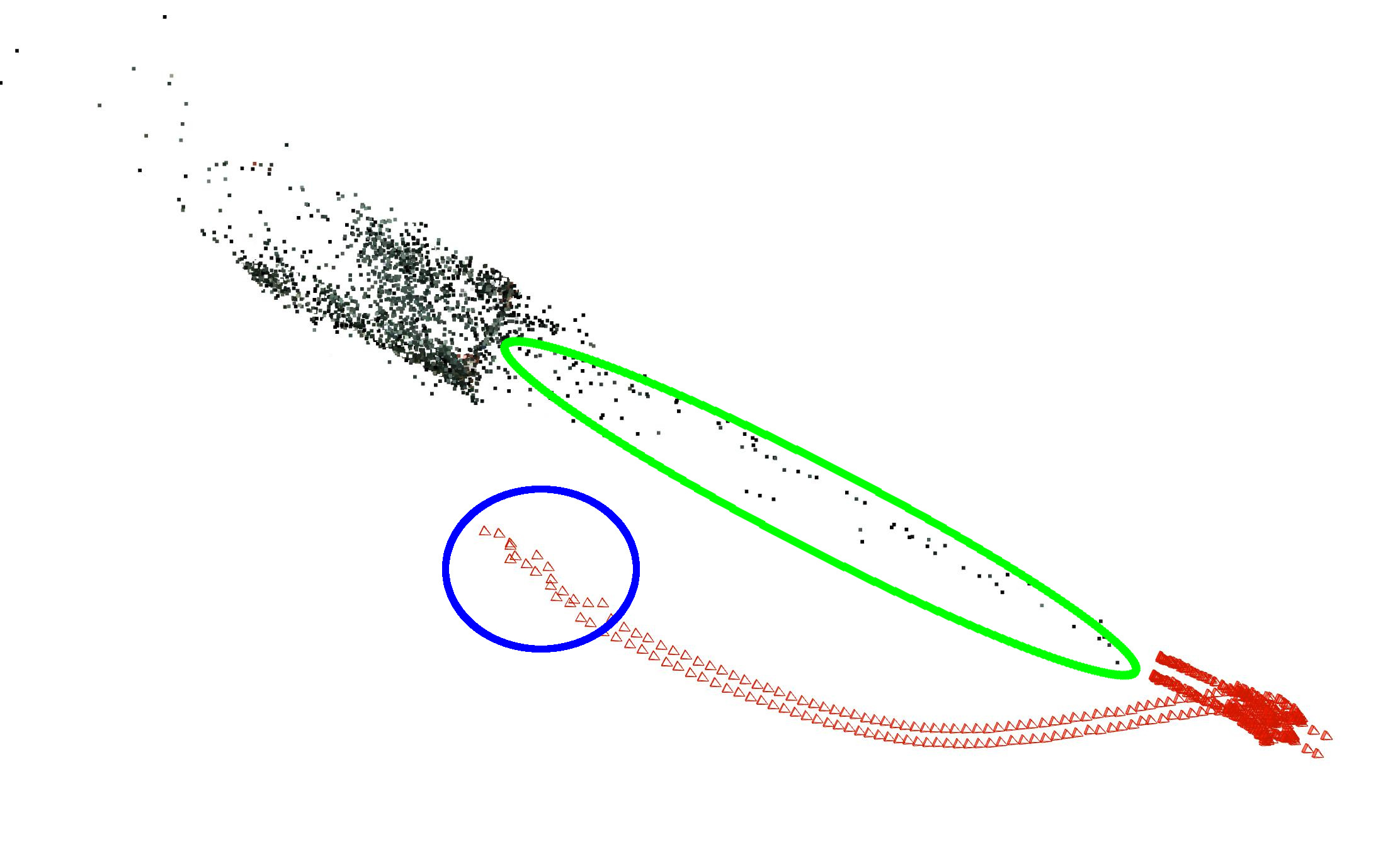}
		\hfill
		\includegraphics[width=\subfigureWidthTwoColumnsInSingleColumn,frame]{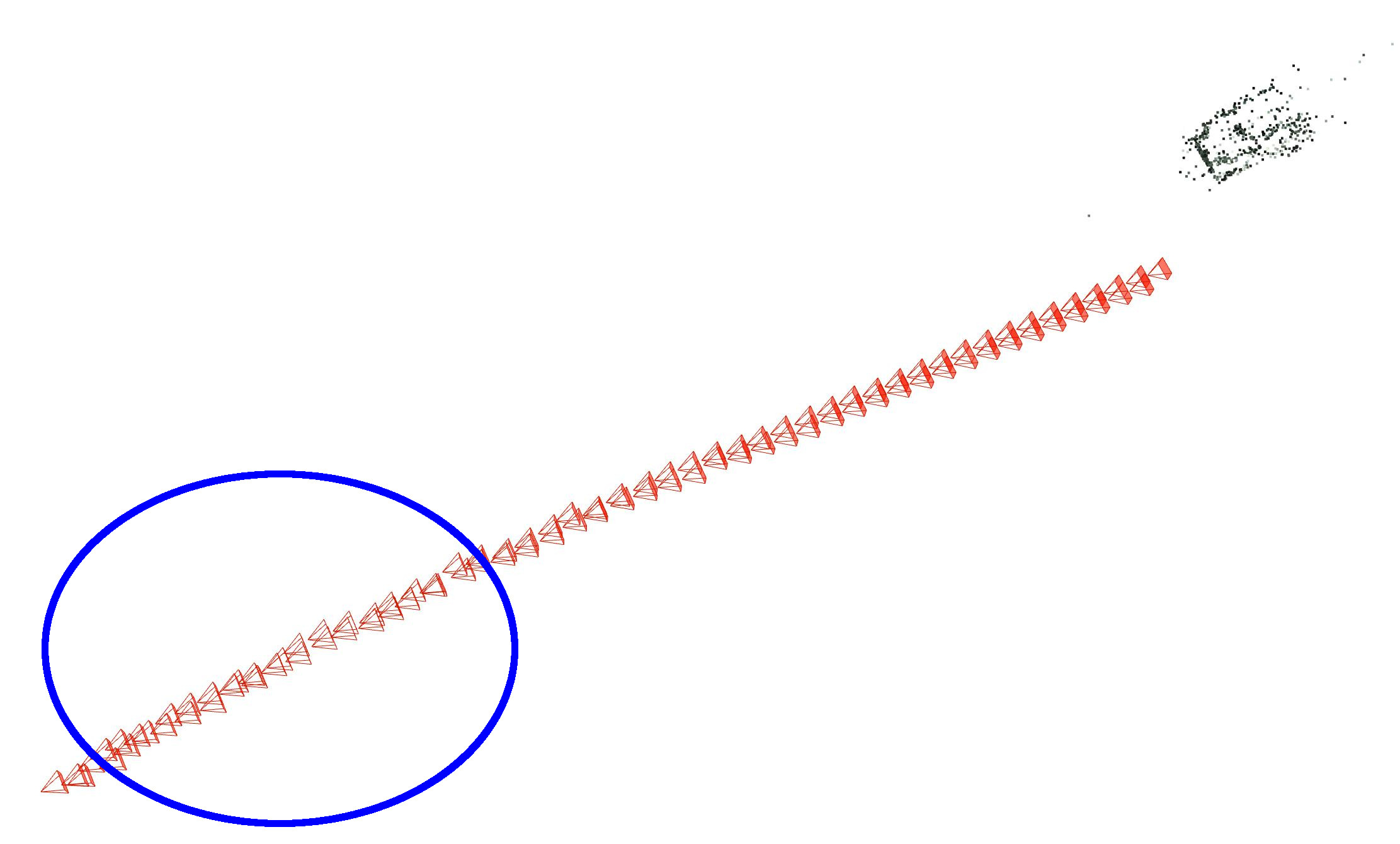}
		%\caption{Initial Object Reconstruction.}
		\label{fig:initial_refinement_comparison1}
	\end{subfigure}
	\hfil
	\begin{subfigure}[t]{0.2\textwidth}
		\includegraphics[width=\subfigureWidthTwoColumnsInSingleColumn,frame]{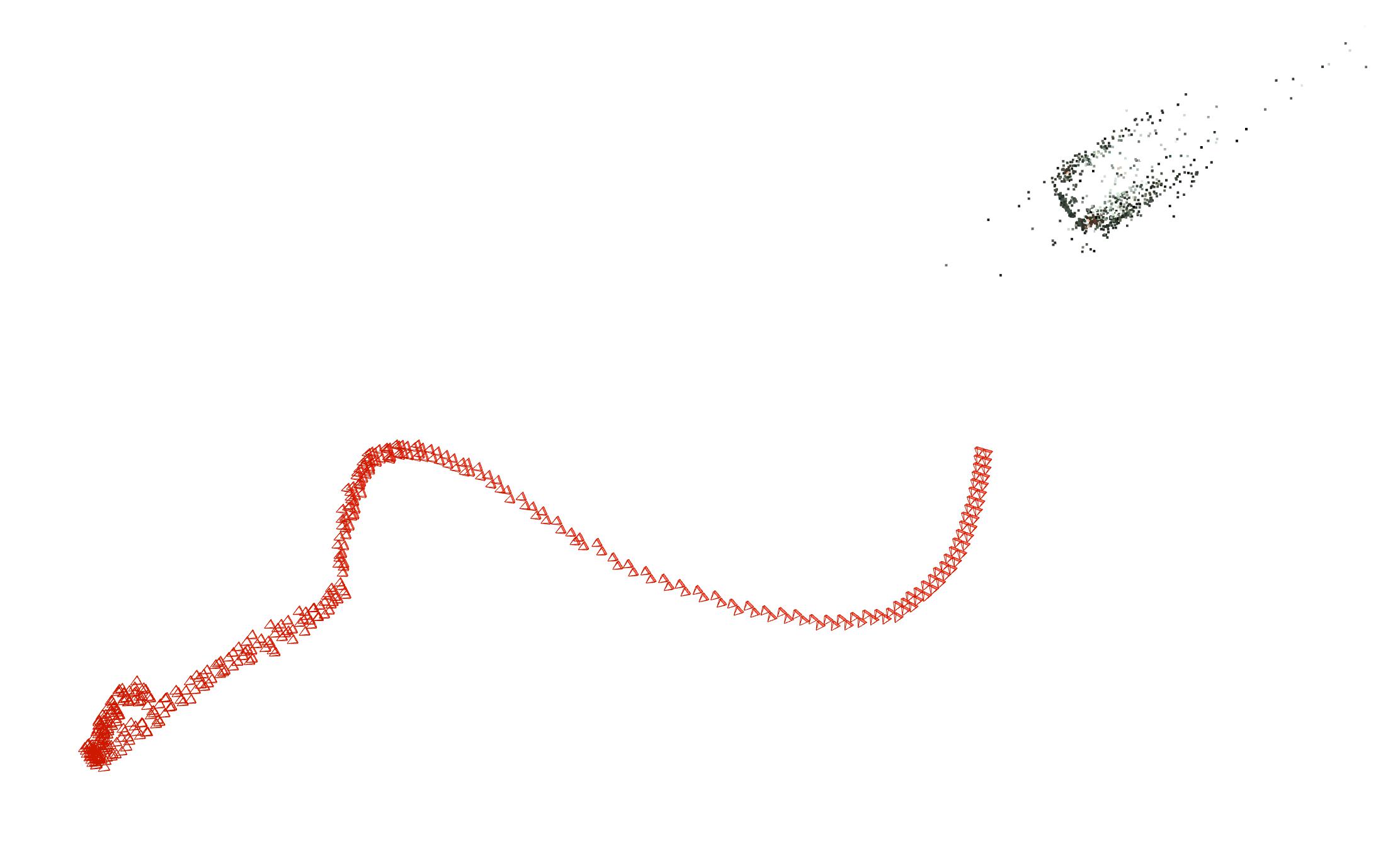}
		\hfill
		\includegraphics[width=\subfigureWidthTwoColumnsInSingleColumn,frame]{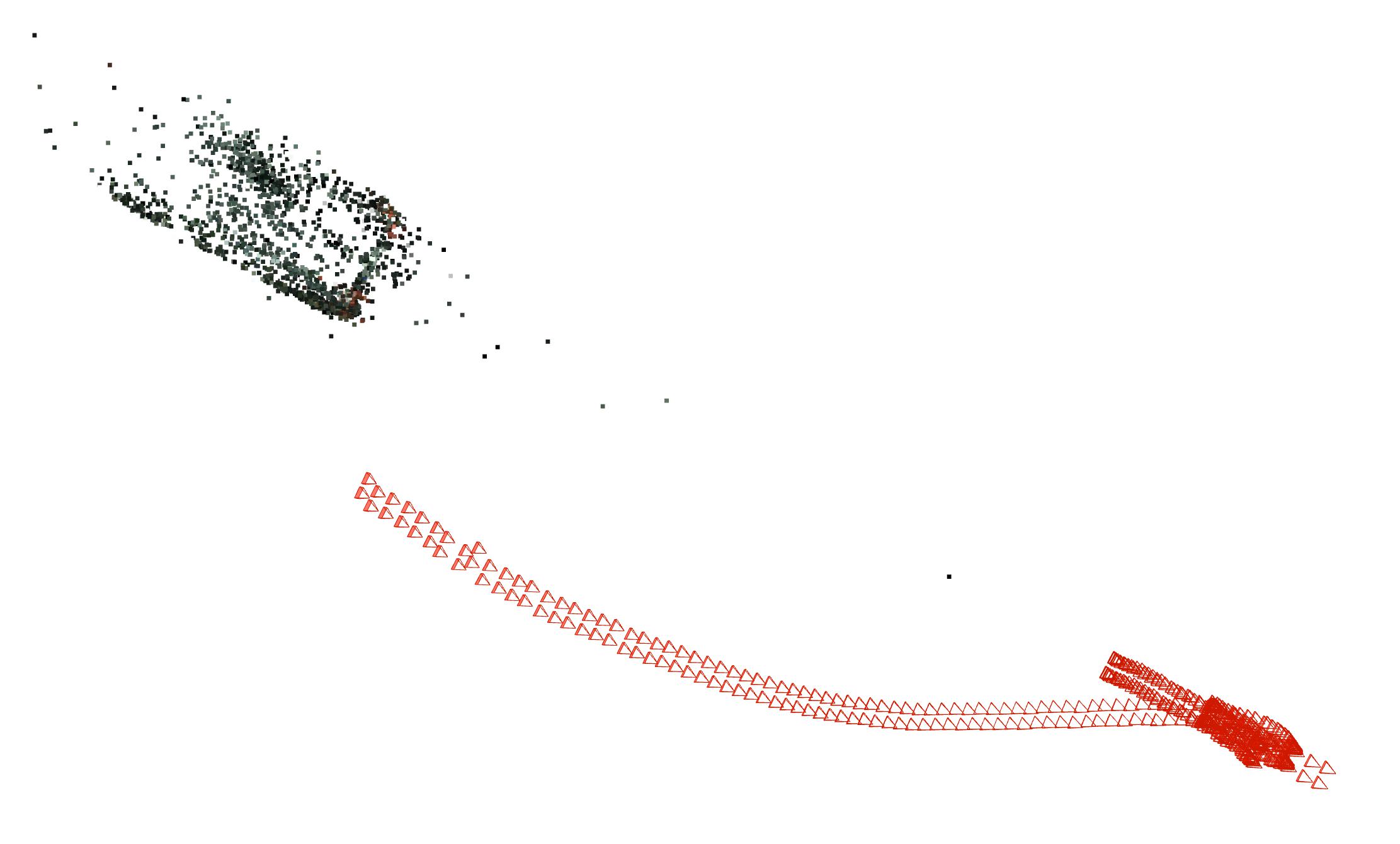}
		\hfill
		\includegraphics[width=\subfigureWidthTwoColumnsInSingleColumn,frame]{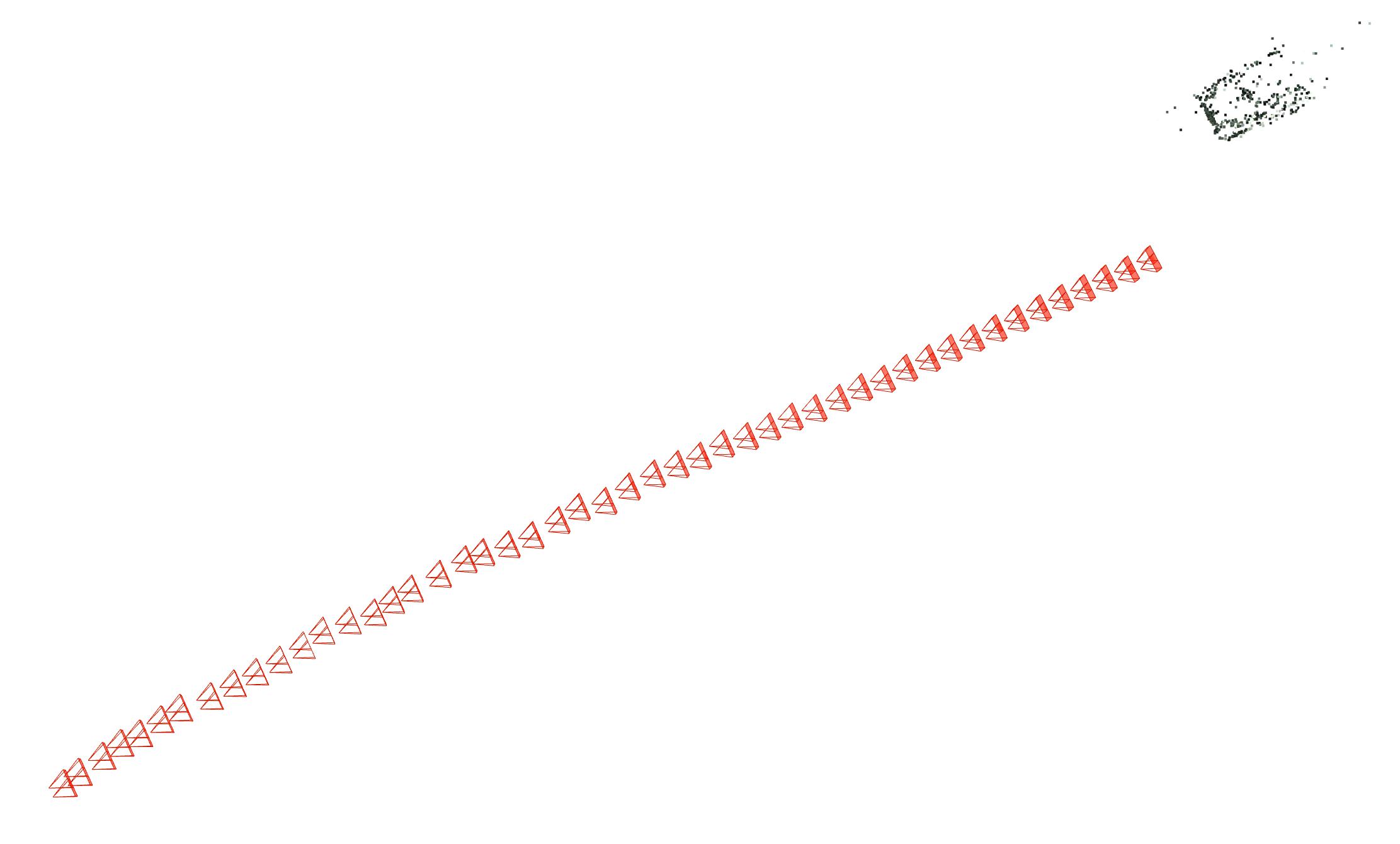}
		%\caption{Object Reconstruction with Stereo Constraints.}
		\label{fig:initial_refinement_comparison2}
	\end{subfigure}

	\caption{Comparison of initial SfM object reconstructions (left column) and corresponding refinements using stereo constraints (right column). The cameras are shown in red. The blue and green circle emphasizes incorrectly registered cameras and triangulated points, respectively.}
\label{methods:registration_error_example}
\end{figure}

%% file: experiments_trajectory_reconstruction_qual_figure.tex
%\newlength{\includegraficsHeightFourColumns} 
%\setlength\includegraficsHeightFourColumns{\includegraficsWidthFourColumns}

% https://tex.stackexchange.com/questions/7219/how-to-vertically-center-two-images-next-to-each-other?utm_medium=organic&utm_source=google_rich_qa&utm_campaign=google_rich_qa

\pgfmathsetmacro{\verticalAlignmentFactor}{0.33}

\begin{figure*}
	% use the subfloat package (not subfigure or subfig)
	\begin{subfigure}[b]{\textwidth}
		\includegraphics[width=\subfigureWidthFourColumns,frame]{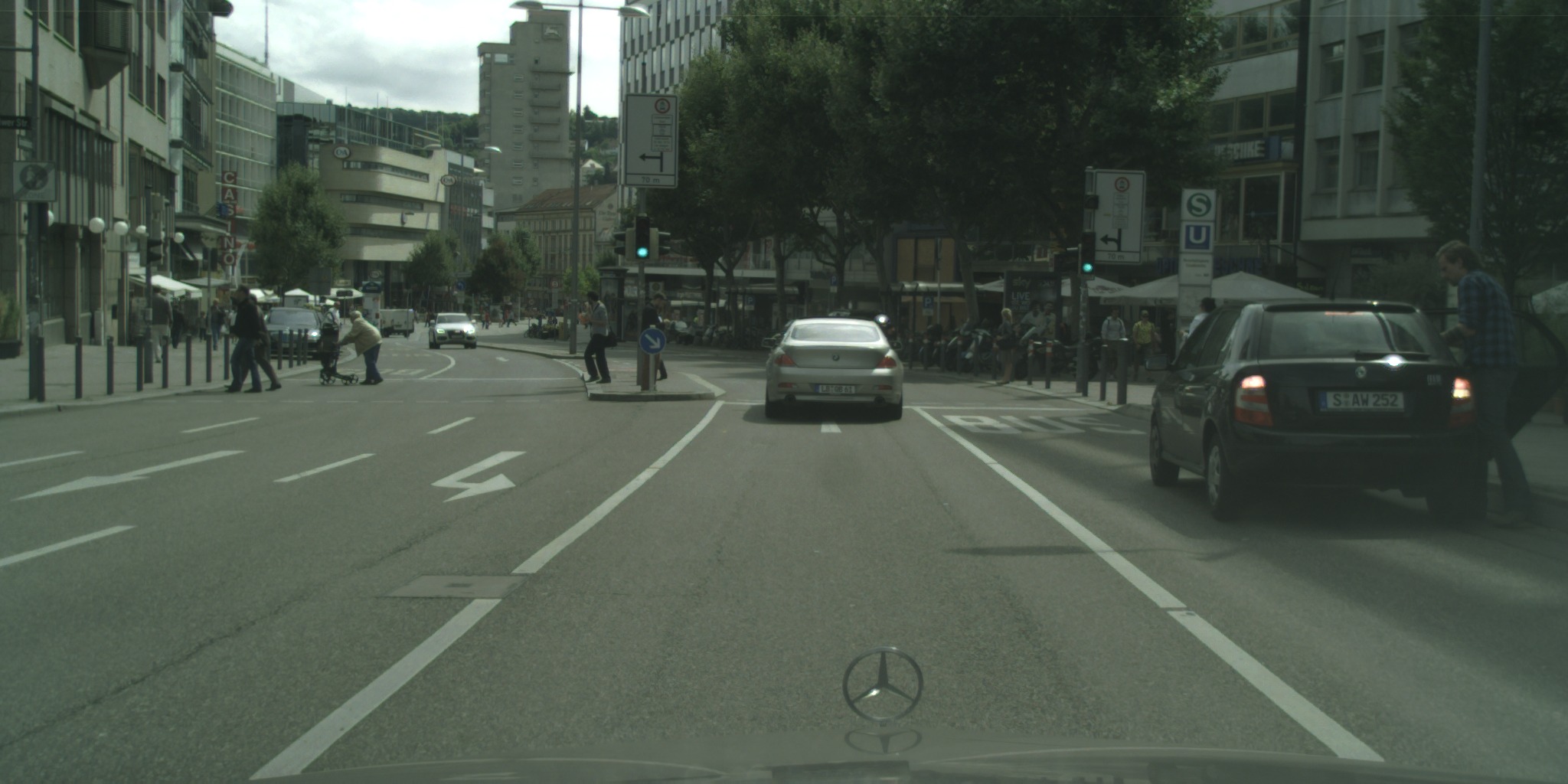}
		\hfill
		\includegraphics[width=\subfigureWidthFourColumns,frame]{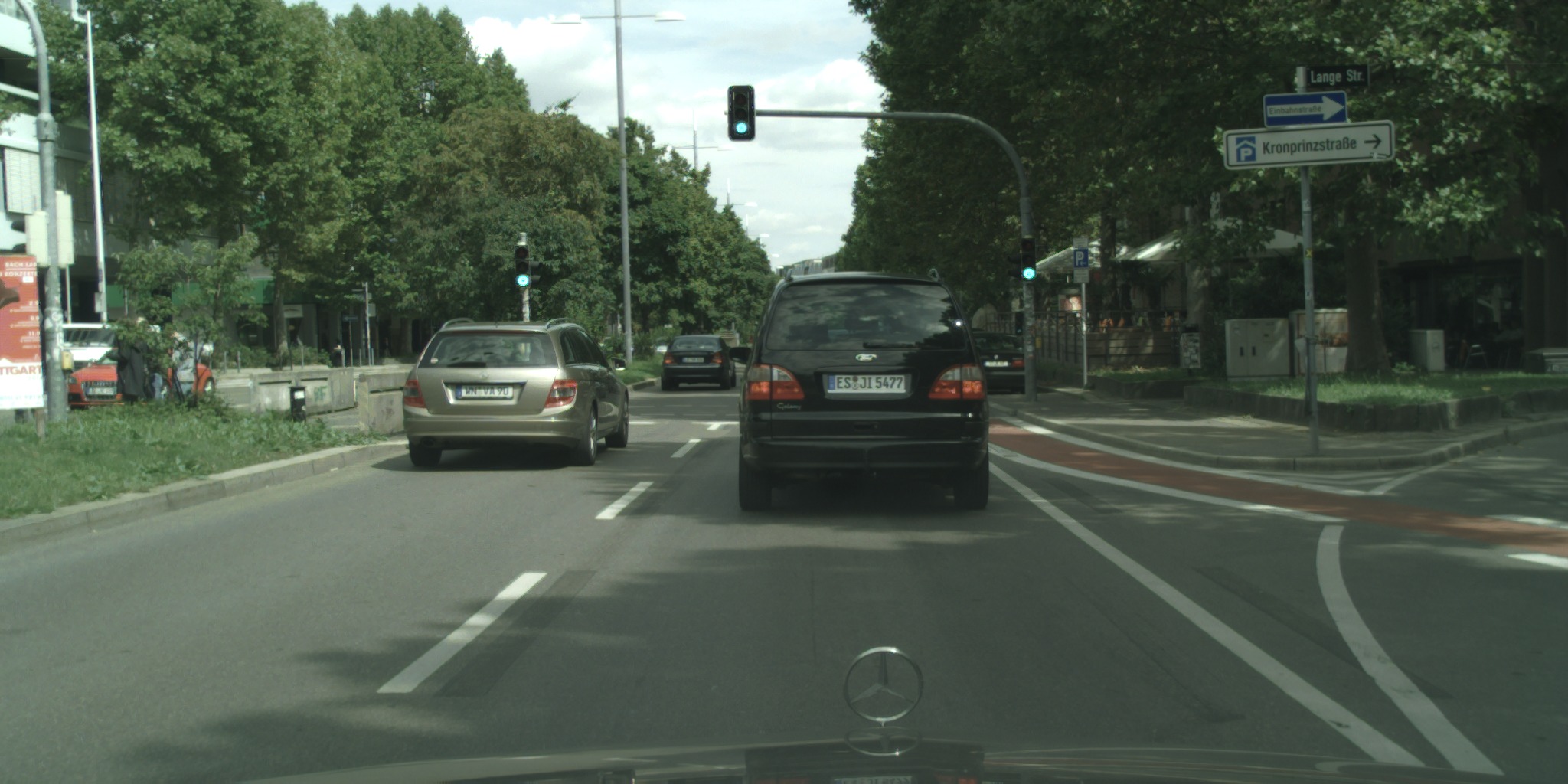}
		\hfill
		\includegraphics[width=\subfigureWidthFourColumns,frame]{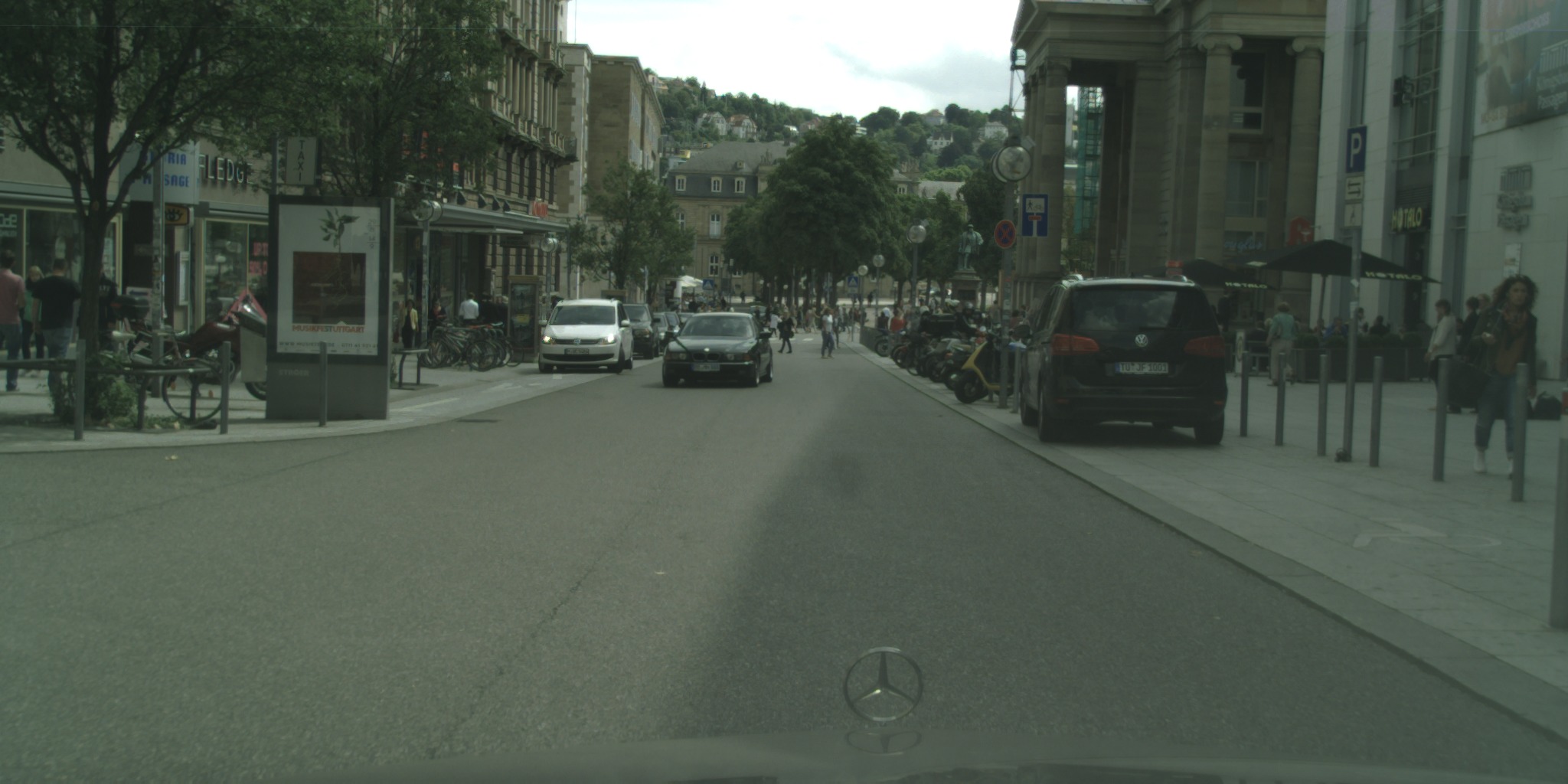}
		\hfill
		\raisebox{\verticalAlignmentFactor\height}{\includegraphics[width=\subfigureWidthFourColumns,frame]{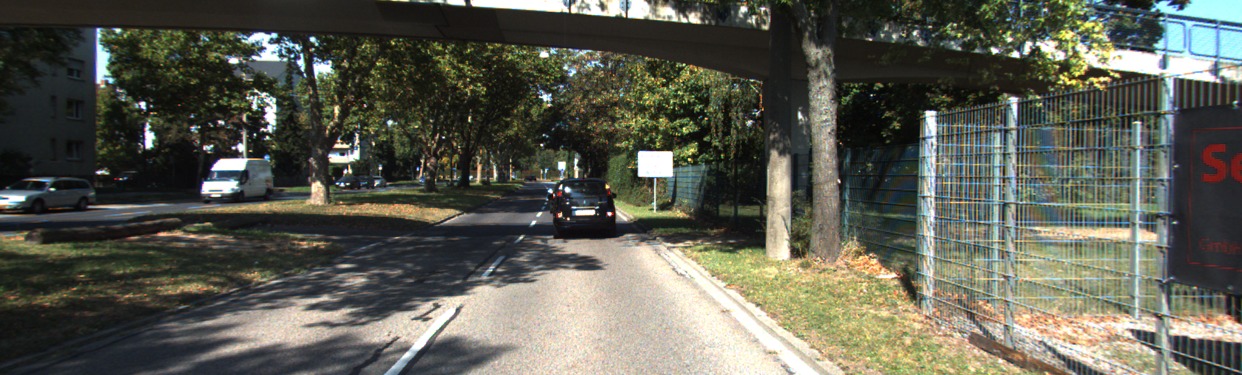}}
		\caption{Left Input Frame.}
		\label{fig:qual_input_frame_0}
	\end{subfigure}
	\hfill
	\begin{subfigure}[t]{\textwidth}
		\includegraphics[width = \subfigureWidthFourColumns,frame]{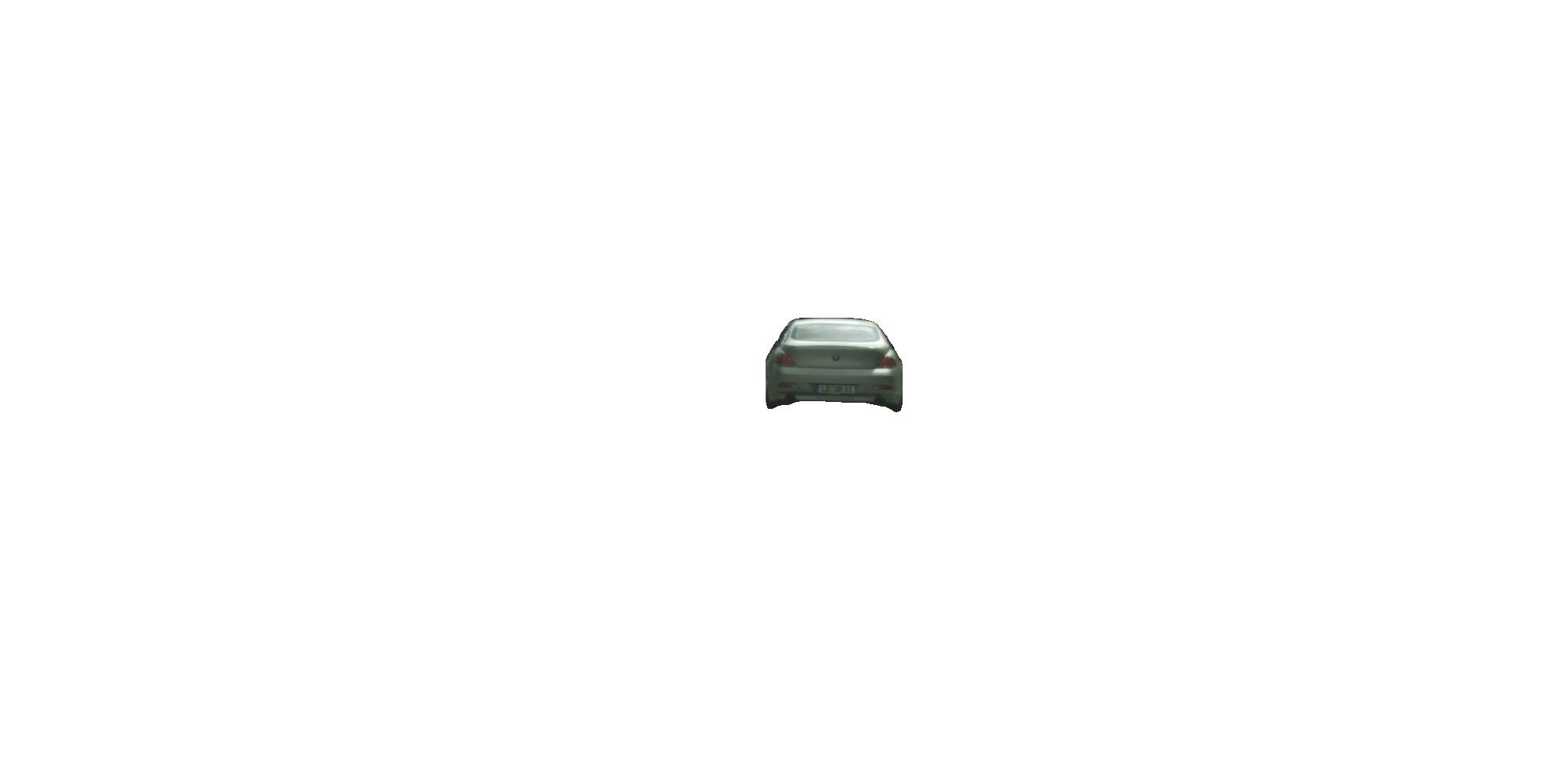}
		\hfill
		\includegraphics[width = \subfigureWidthFourColumns,frame]{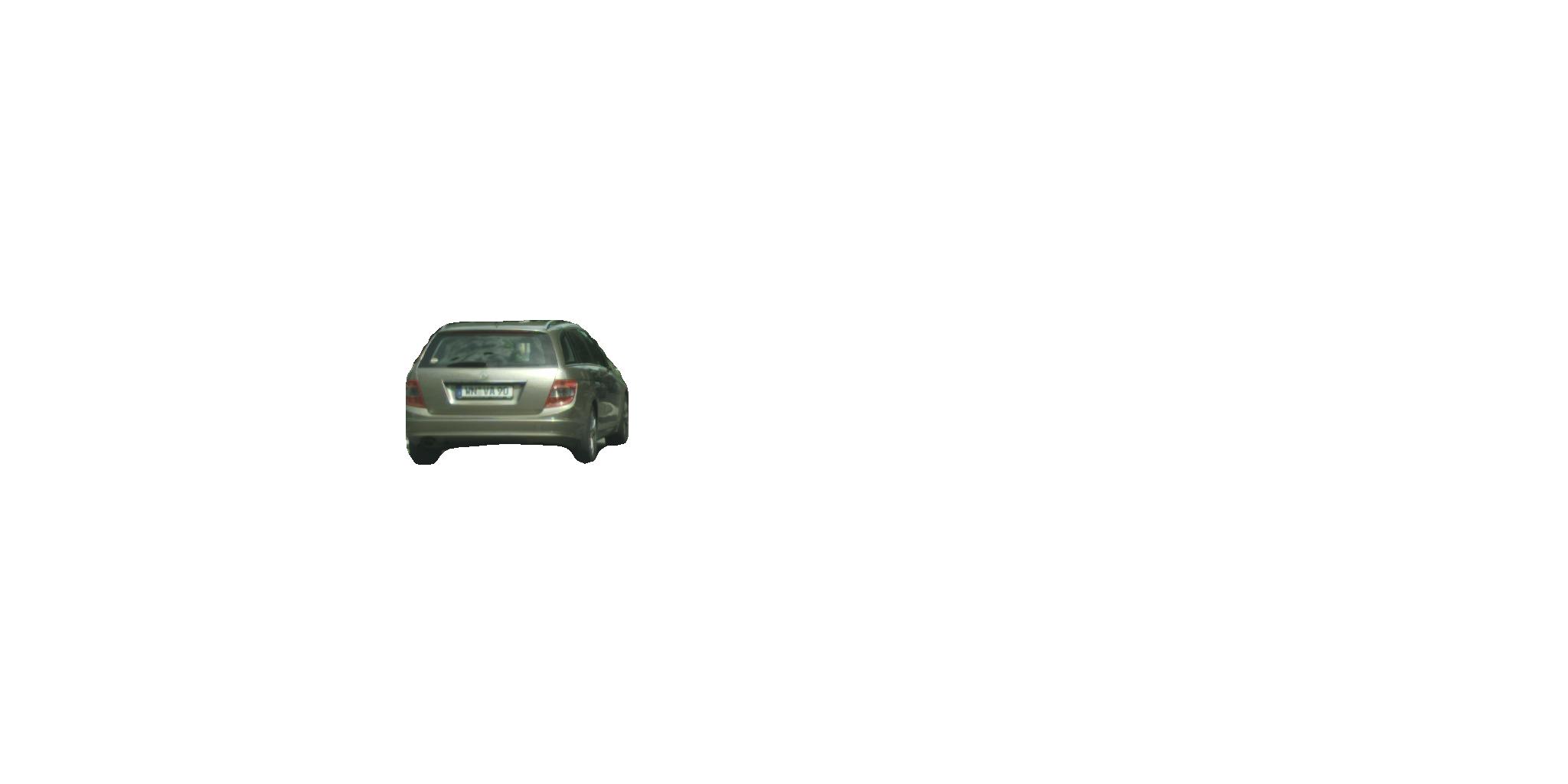}
		\hfill
		\includegraphics[width = \subfigureWidthFourColumns,frame]{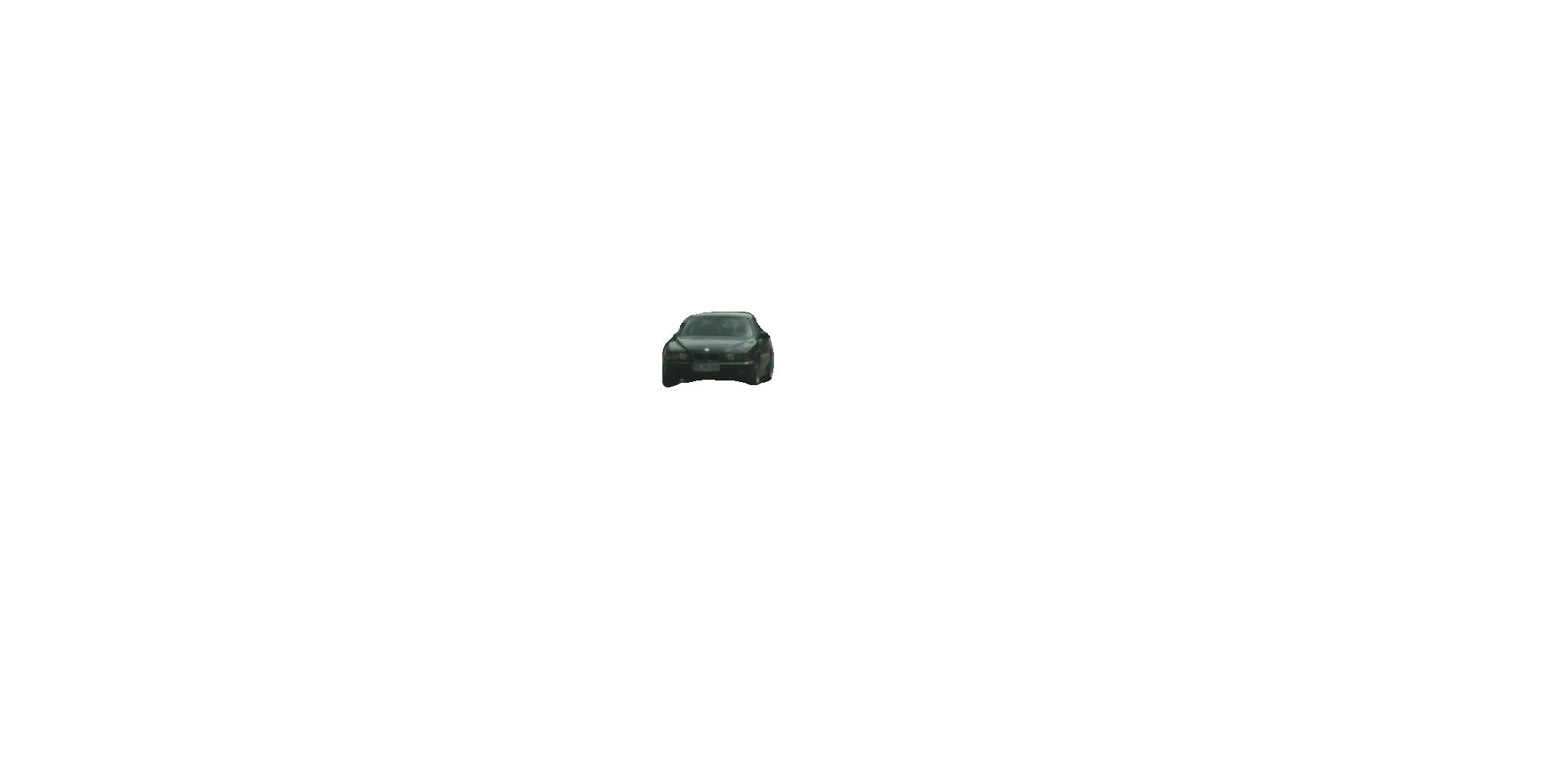}
		\hfill
		\raisebox{\verticalAlignmentFactor\height}{\includegraphics[width = \subfigureWidthFourColumns,frame]{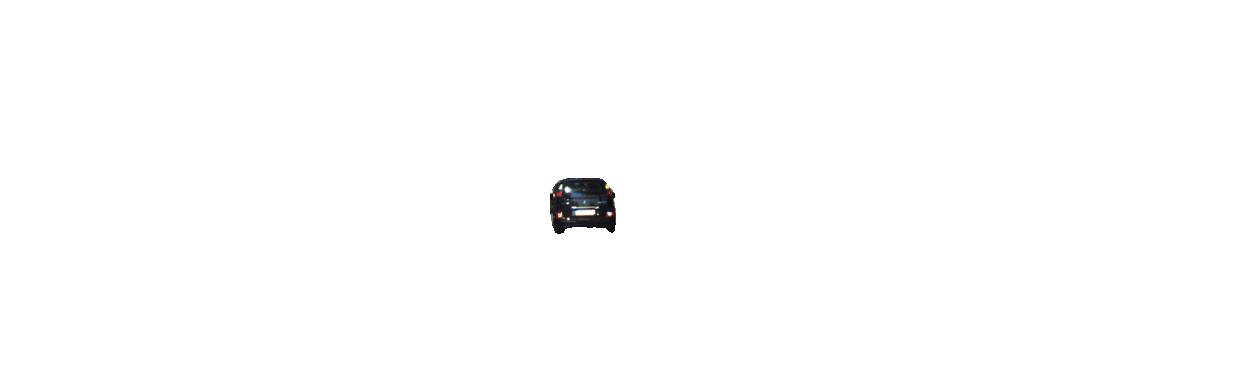}}
		\caption{Left Object Segmentation.}
		\label{fig:qual_obj_seg}
	\end{subfigure}
	\hfill
	\begin{subfigure}[b]{\textwidth}
		\includegraphics[width = \subfigureWidthFourColumns,frame]{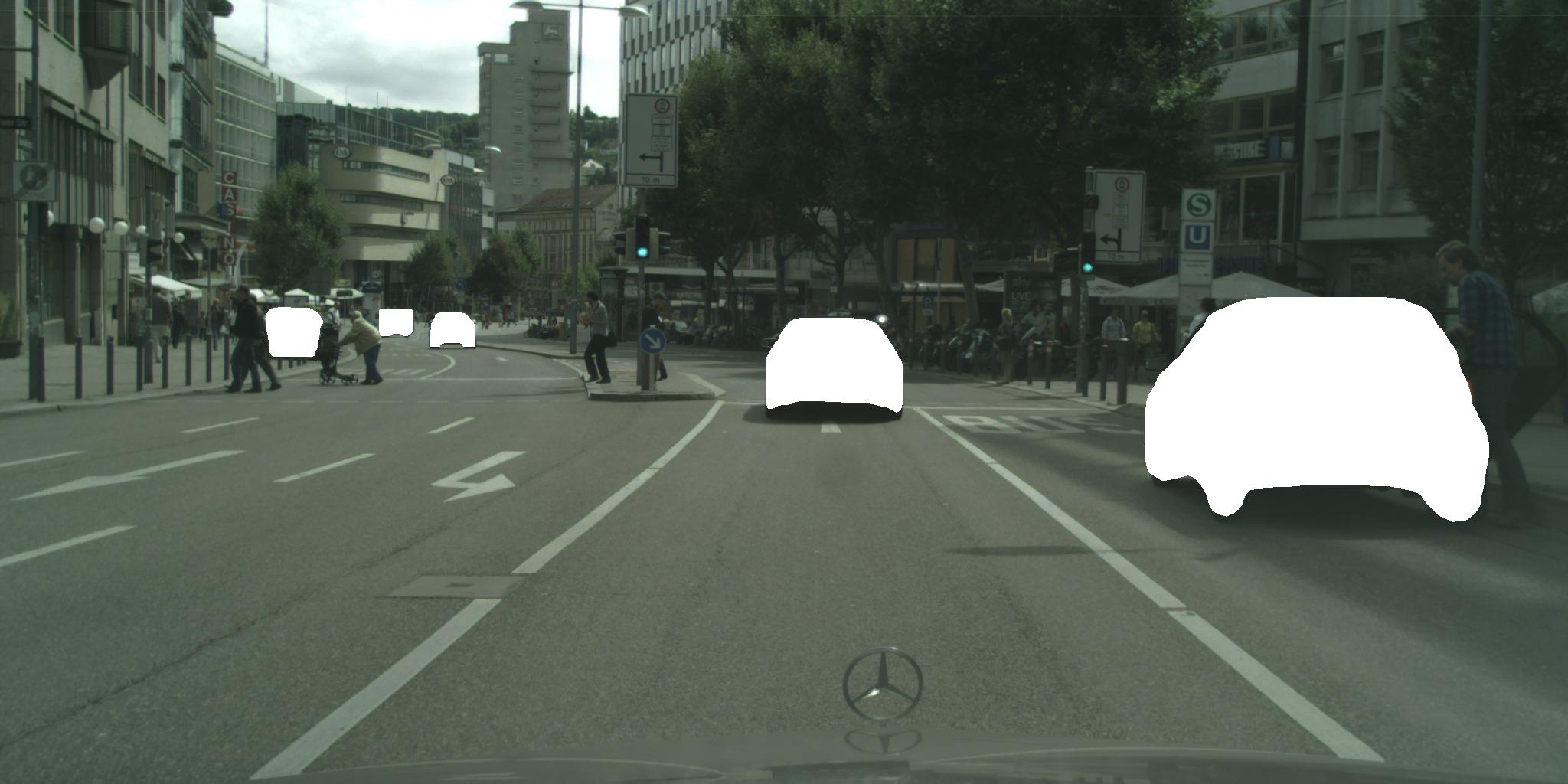}
		\hfill
		\includegraphics[width = \subfigureWidthFourColumns,frame]{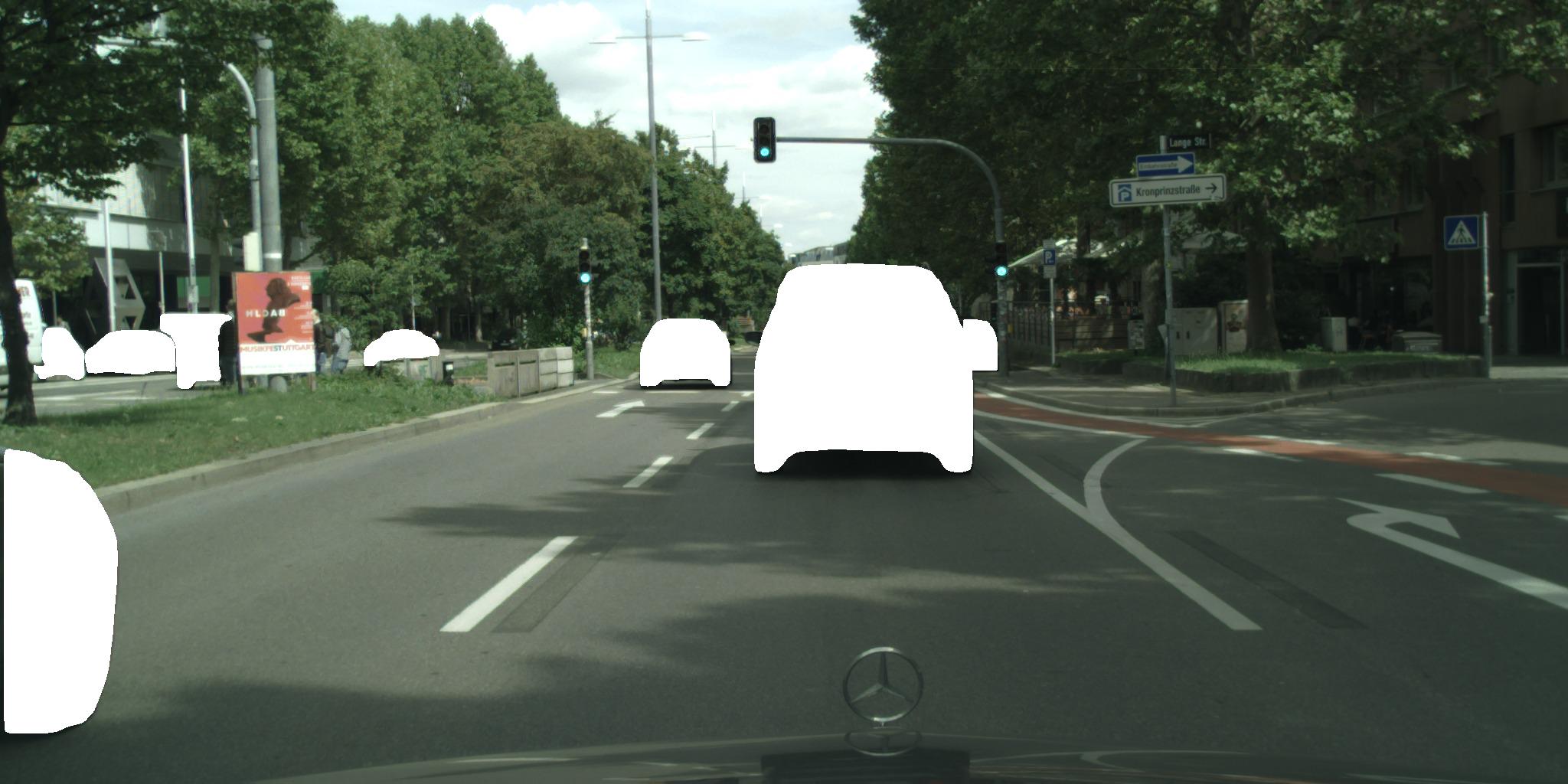}
		\hfill
		\includegraphics[width = \subfigureWidthFourColumns,frame]{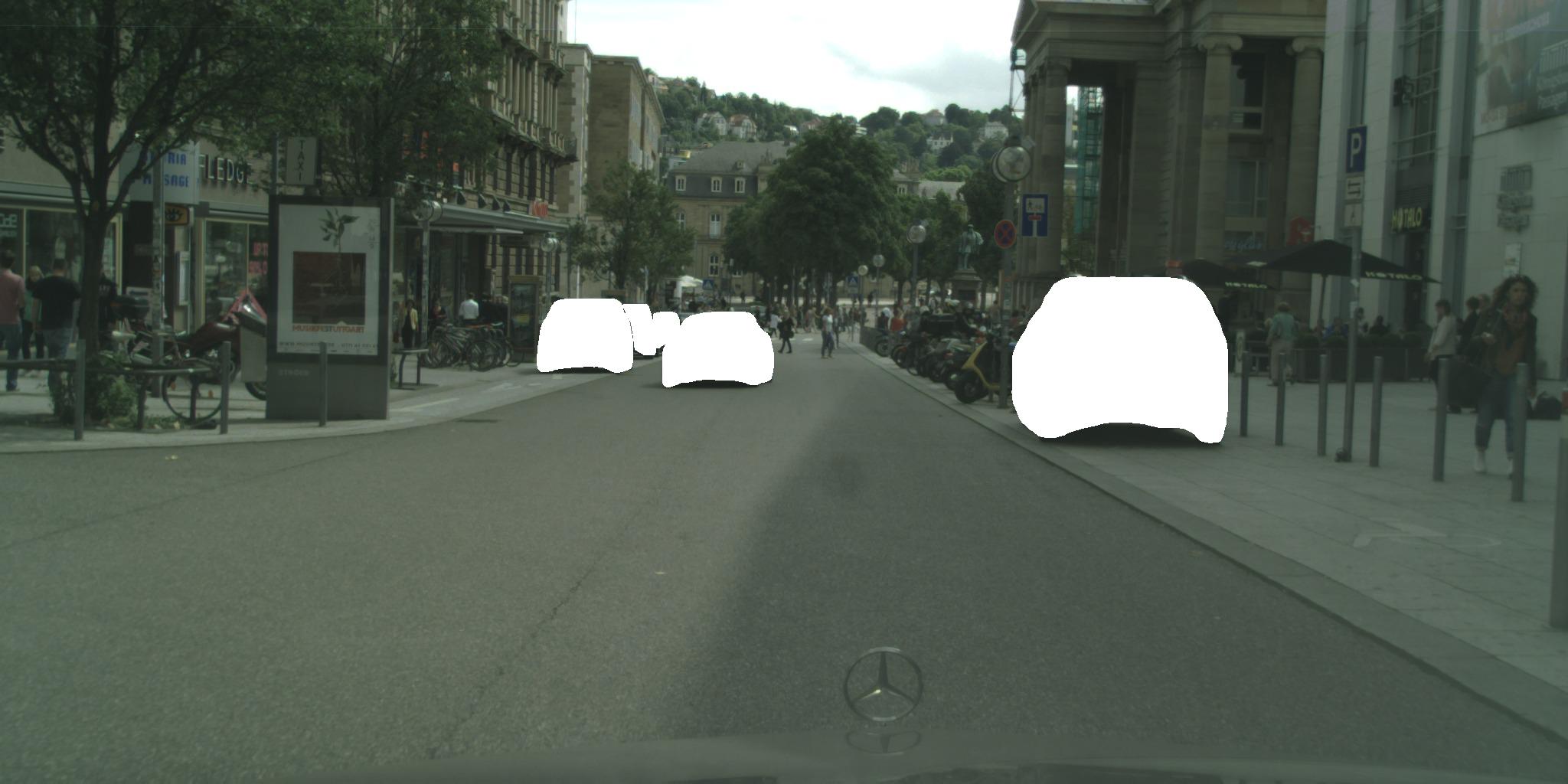}
		\hfill
		\raisebox{\verticalAlignmentFactor\height}{\includegraphics[width = \subfigureWidthFourColumns,frame]{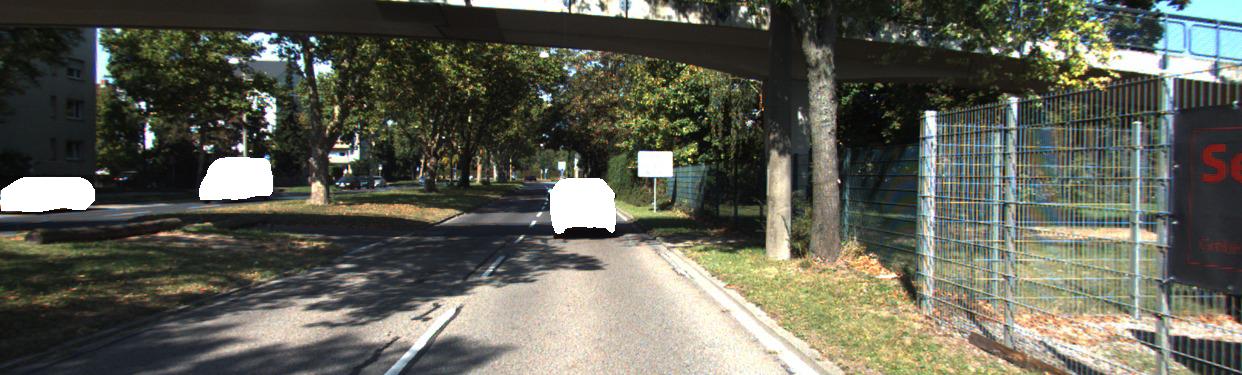}}
		\caption{Left Background Segmentation.}
		\label{fig:qual_back_seg}
	\end{subfigure}
	\hfill
	\begin{subfigure}[t]{\textwidth}
		\includegraphics[width = \subfigureWidthFourColumns,frame]{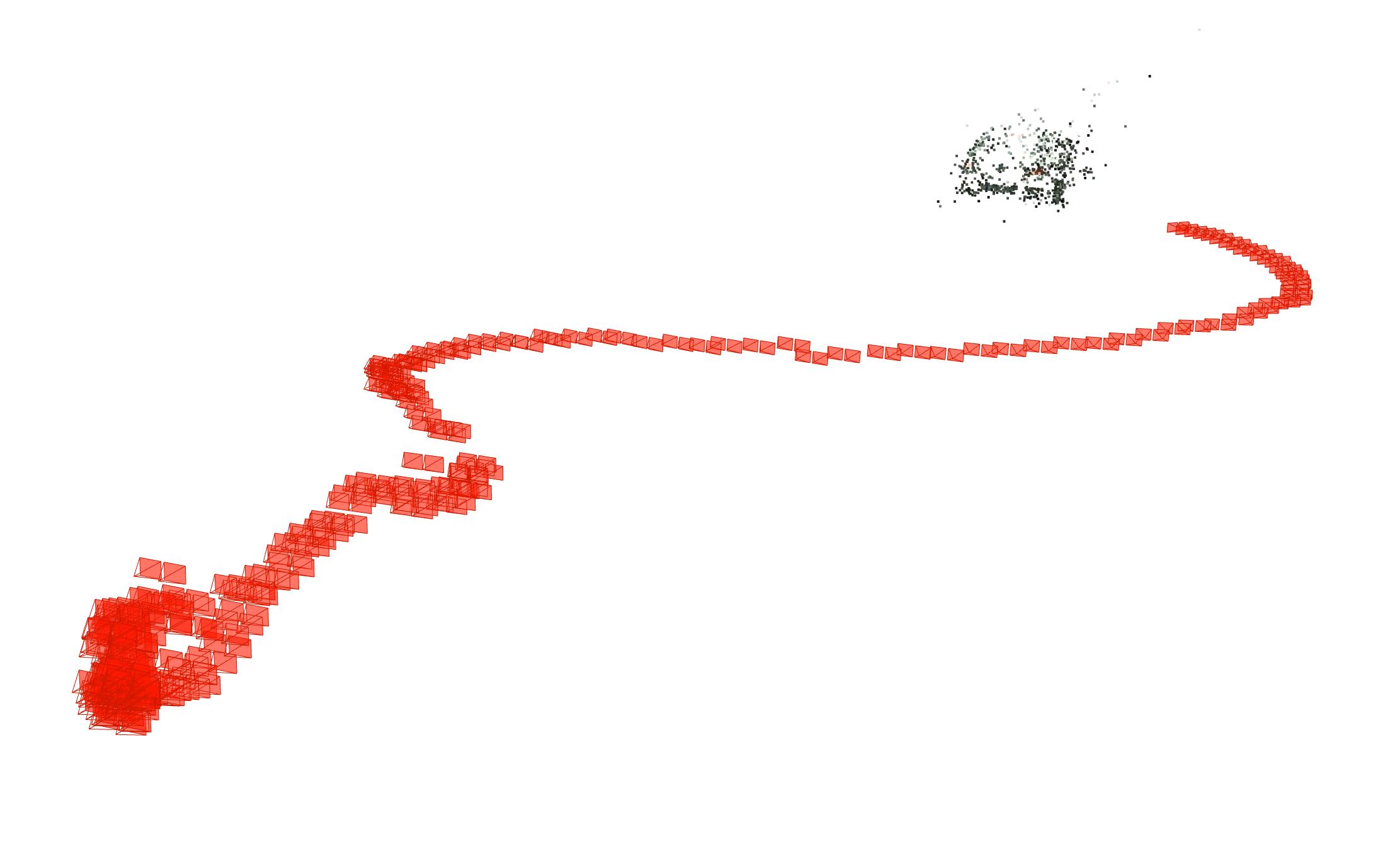}
		\hfill
		\includegraphics[width = \subfigureWidthFourColumns,frame]{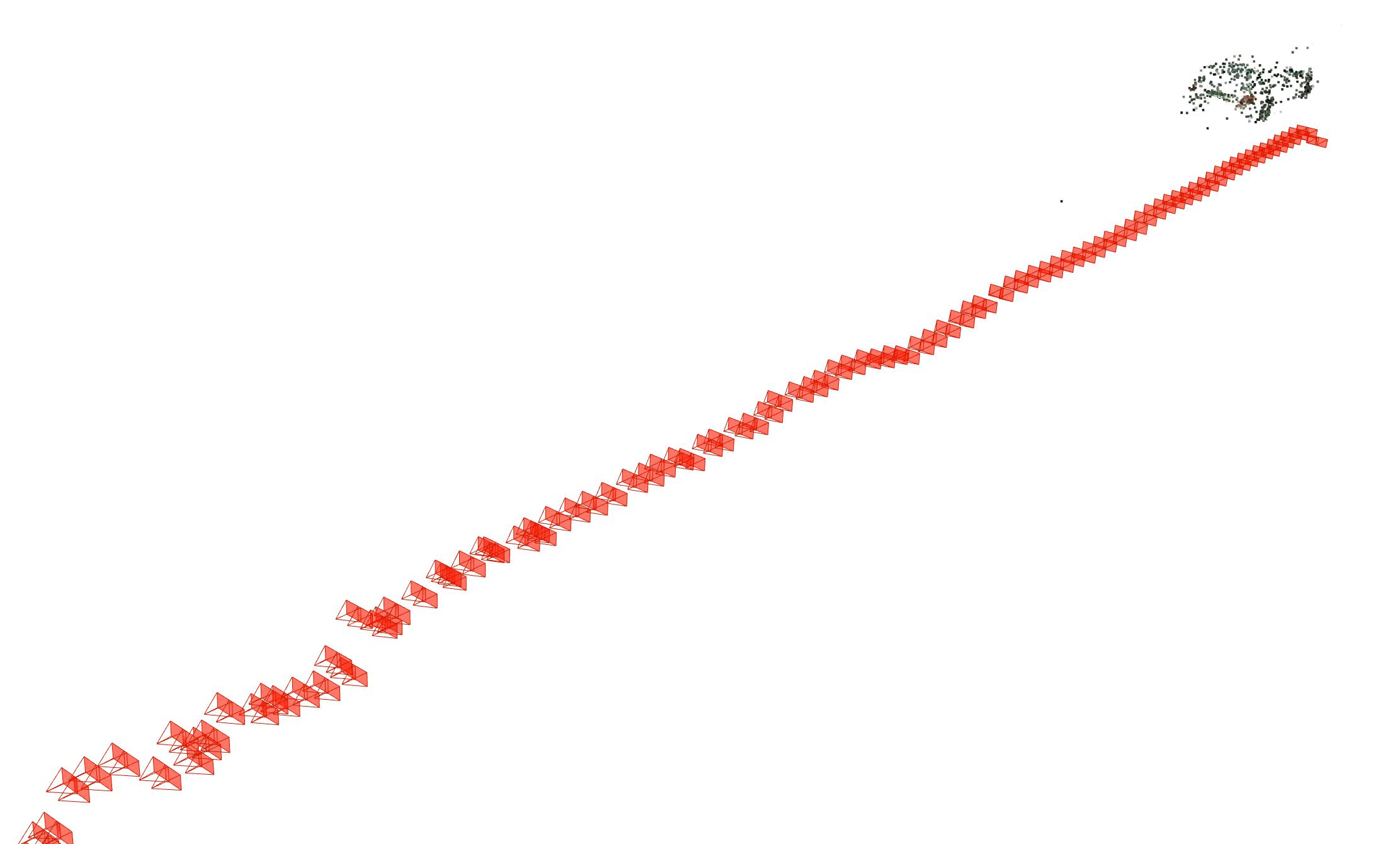}
		\hfill
		\includegraphics[width = \subfigureWidthFourColumns,frame]{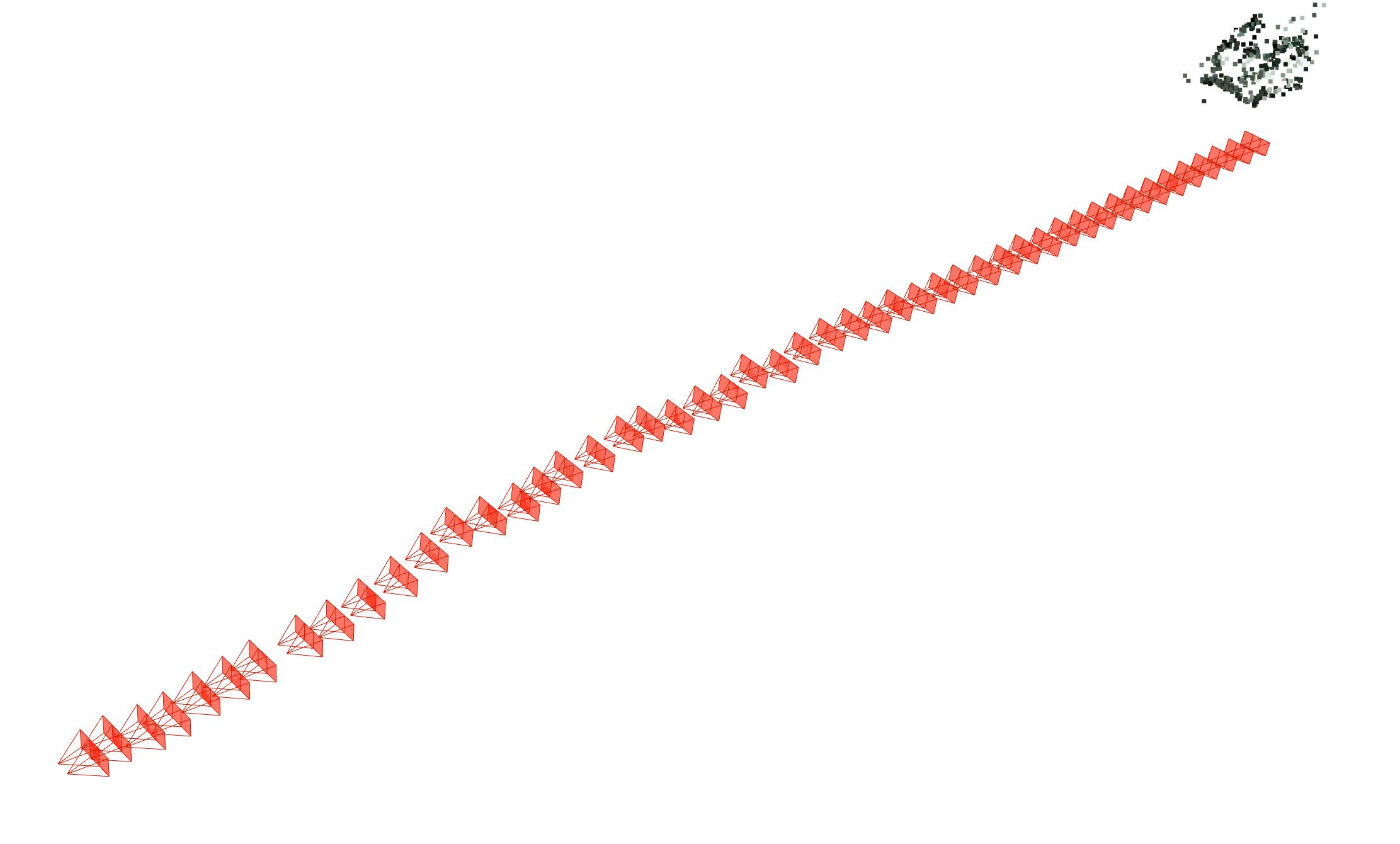}
		\hfill
		\includegraphics[width = \subfigureWidthFourColumns,frame]{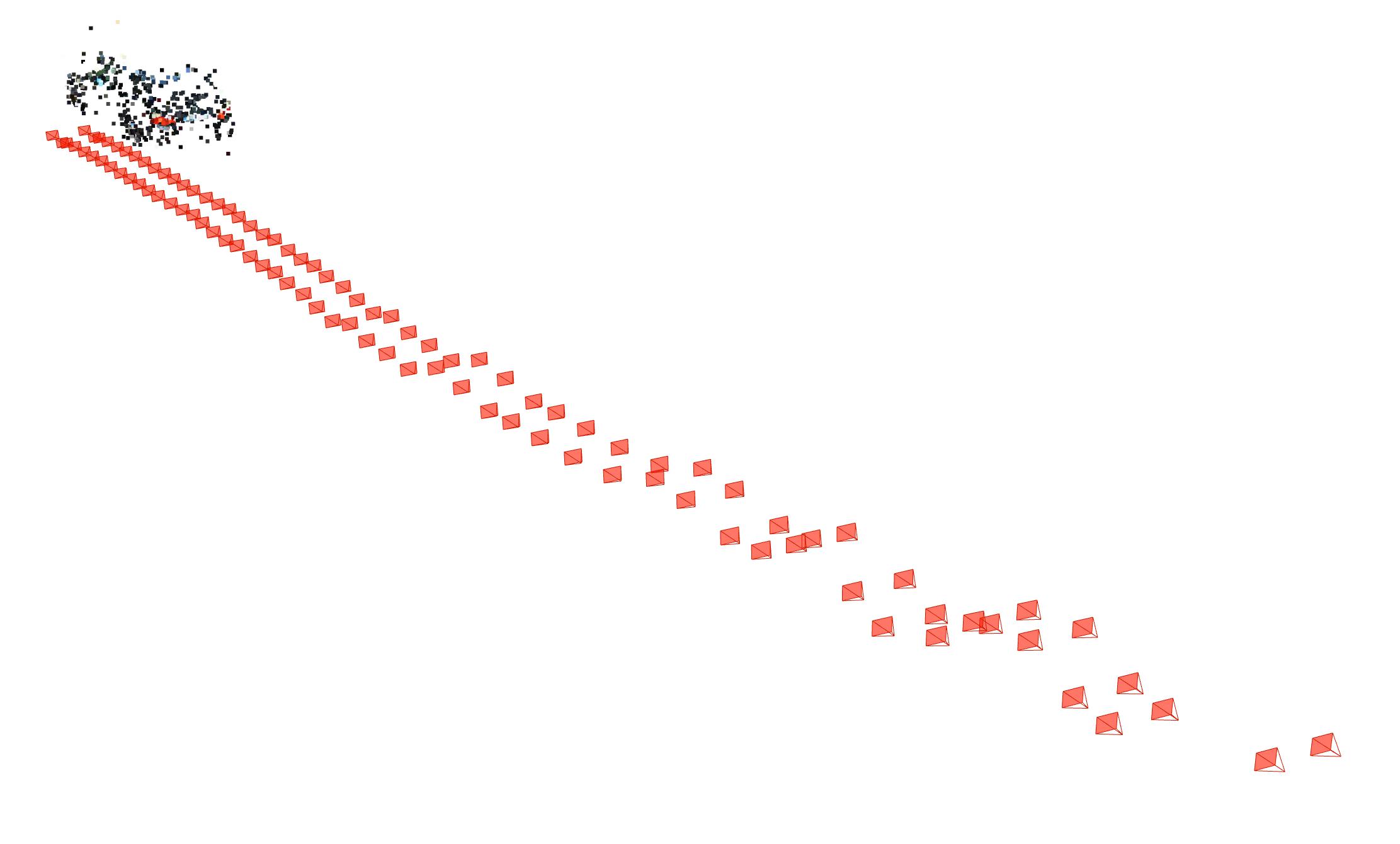}
		\caption{Object Reconstruction.}
		\label{fig:qual_object_rec}
	\end{subfigure}
	\hfill
	\begin{subfigure}[t]{\textwidth}
		\includegraphics[width = \subfigureWidthFourColumns,frame]{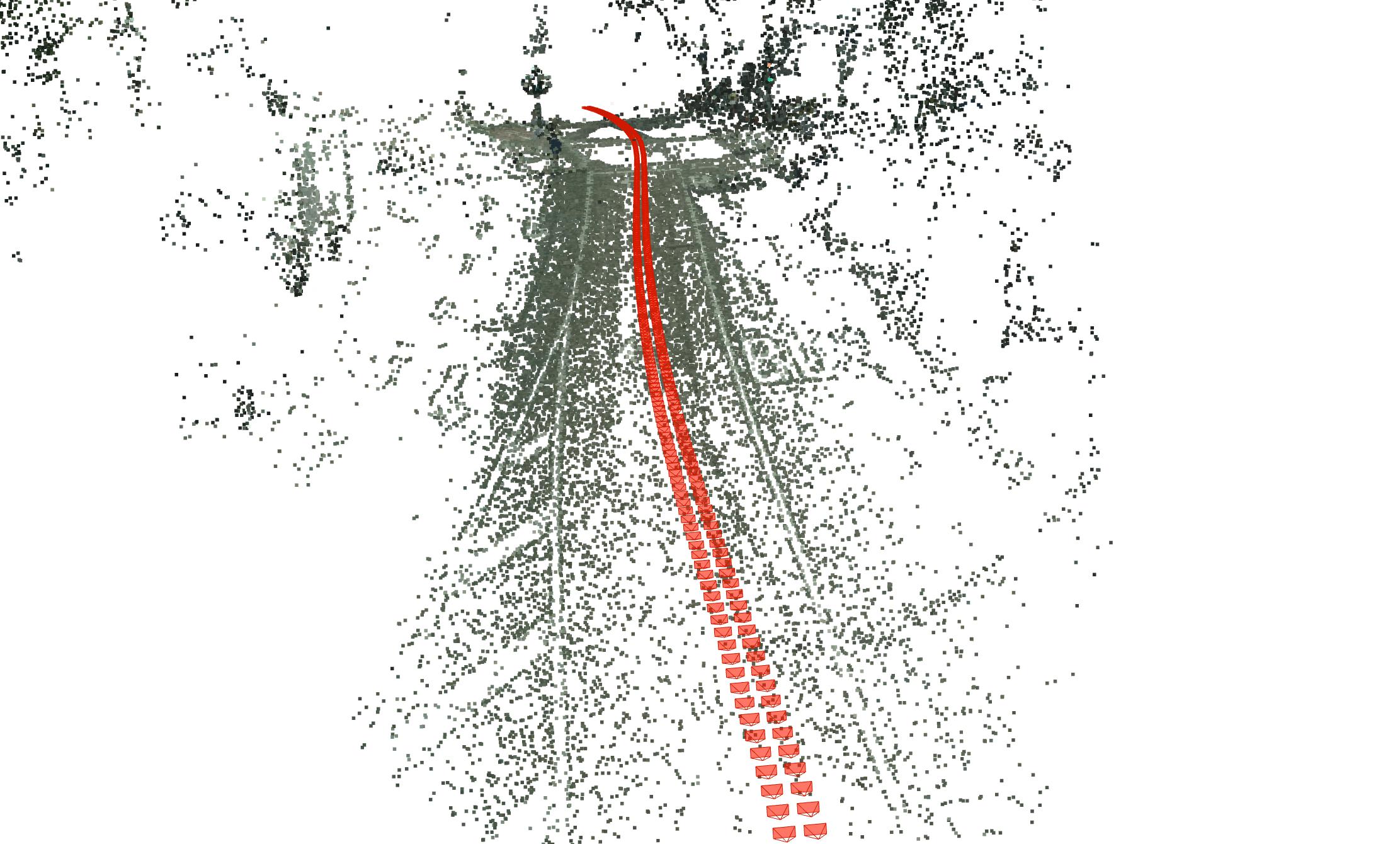}
		\hfill
		\includegraphics[width = \subfigureWidthFourColumns,frame]{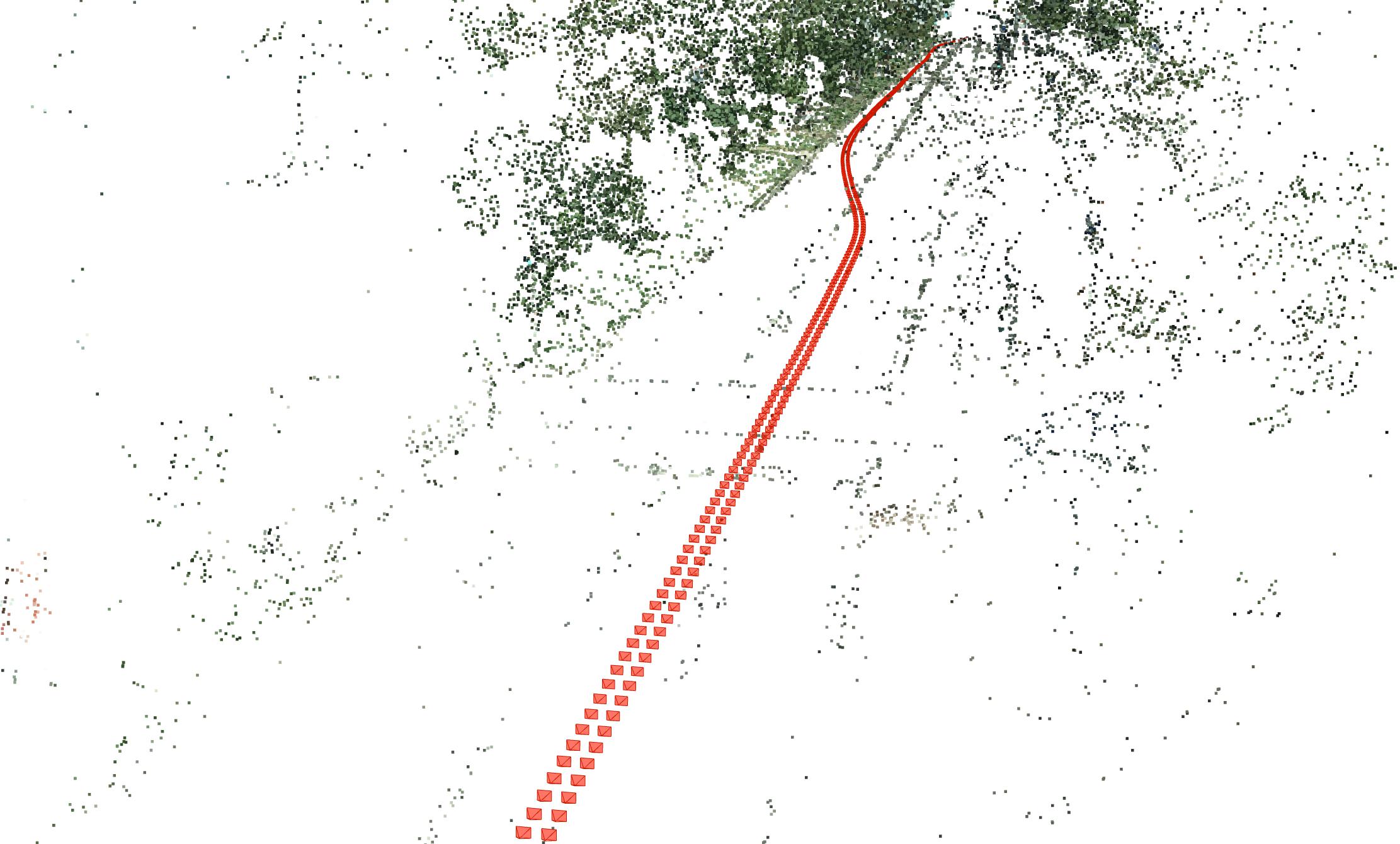}
		\hfill
		\includegraphics[width = \subfigureWidthFourColumns,frame]{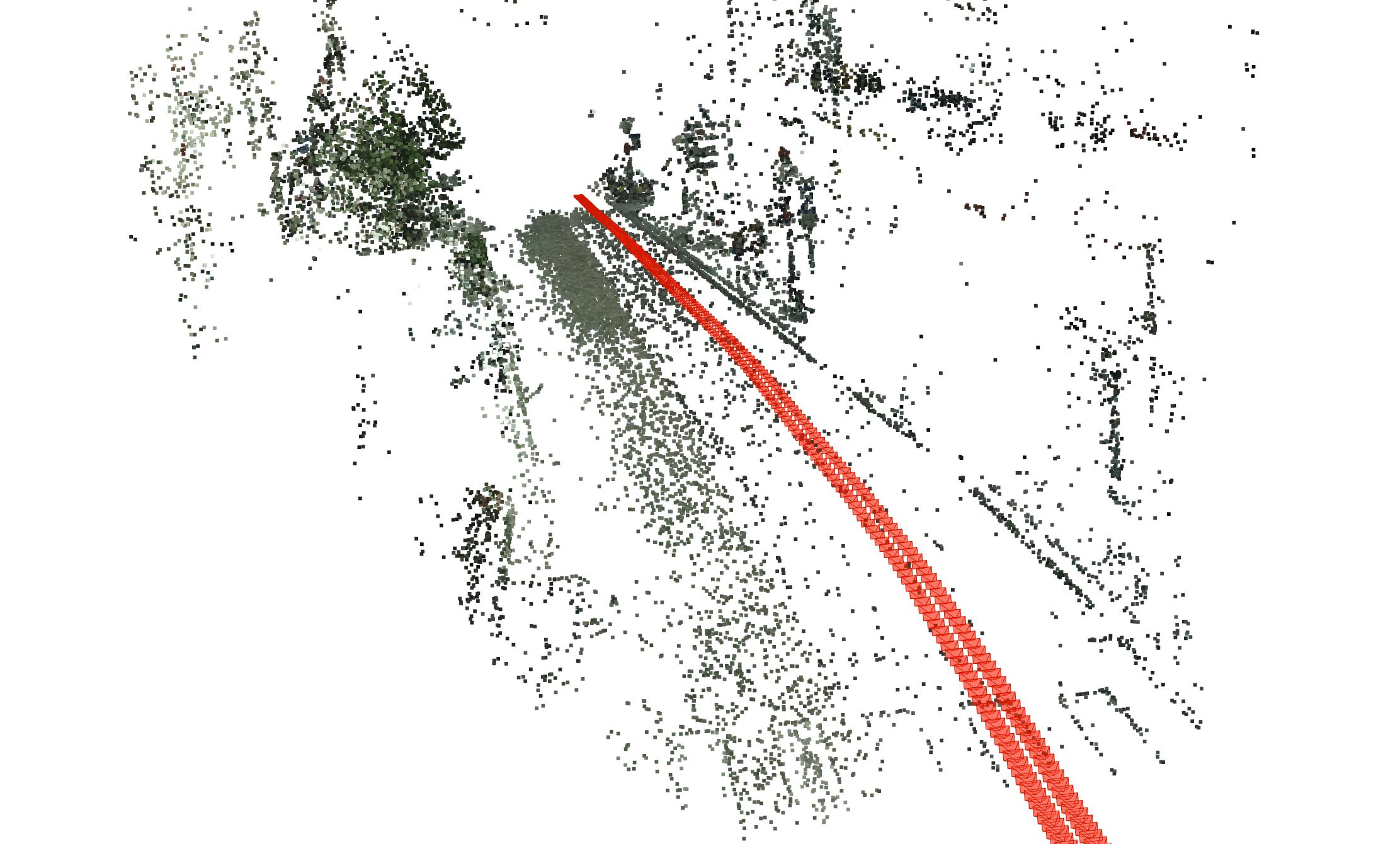}
		\hfill
		\includegraphics[width = \subfigureWidthFourColumns,frame]{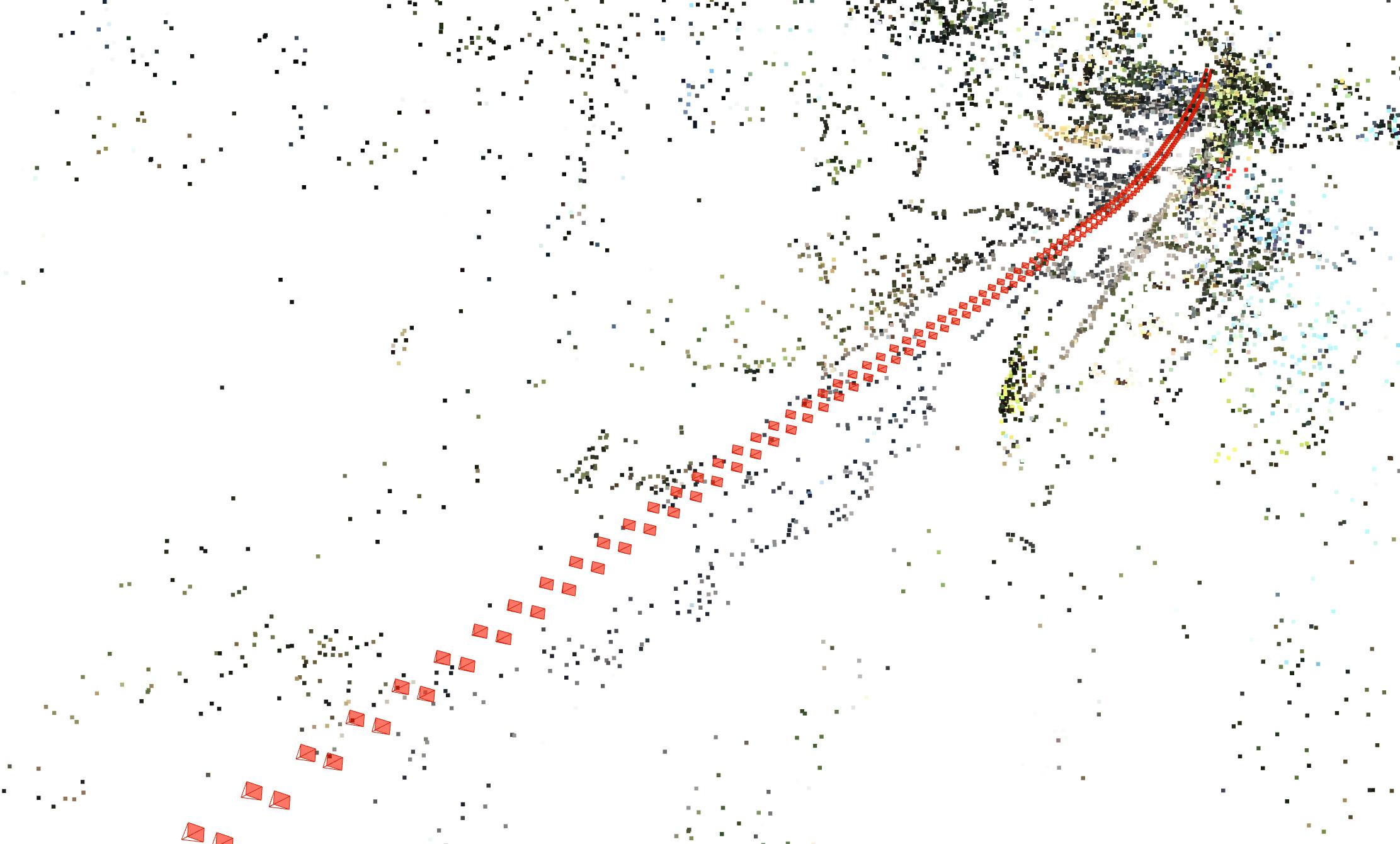}
		\caption{Background Reconstruction.}
		\label{fig:qual_background_rec}
	\end{subfigure}
	\hfill
	\begin{subfigure}[t]{\textwidth}
		\includegraphics[width = \subfigureWidthFourColumns,frame]{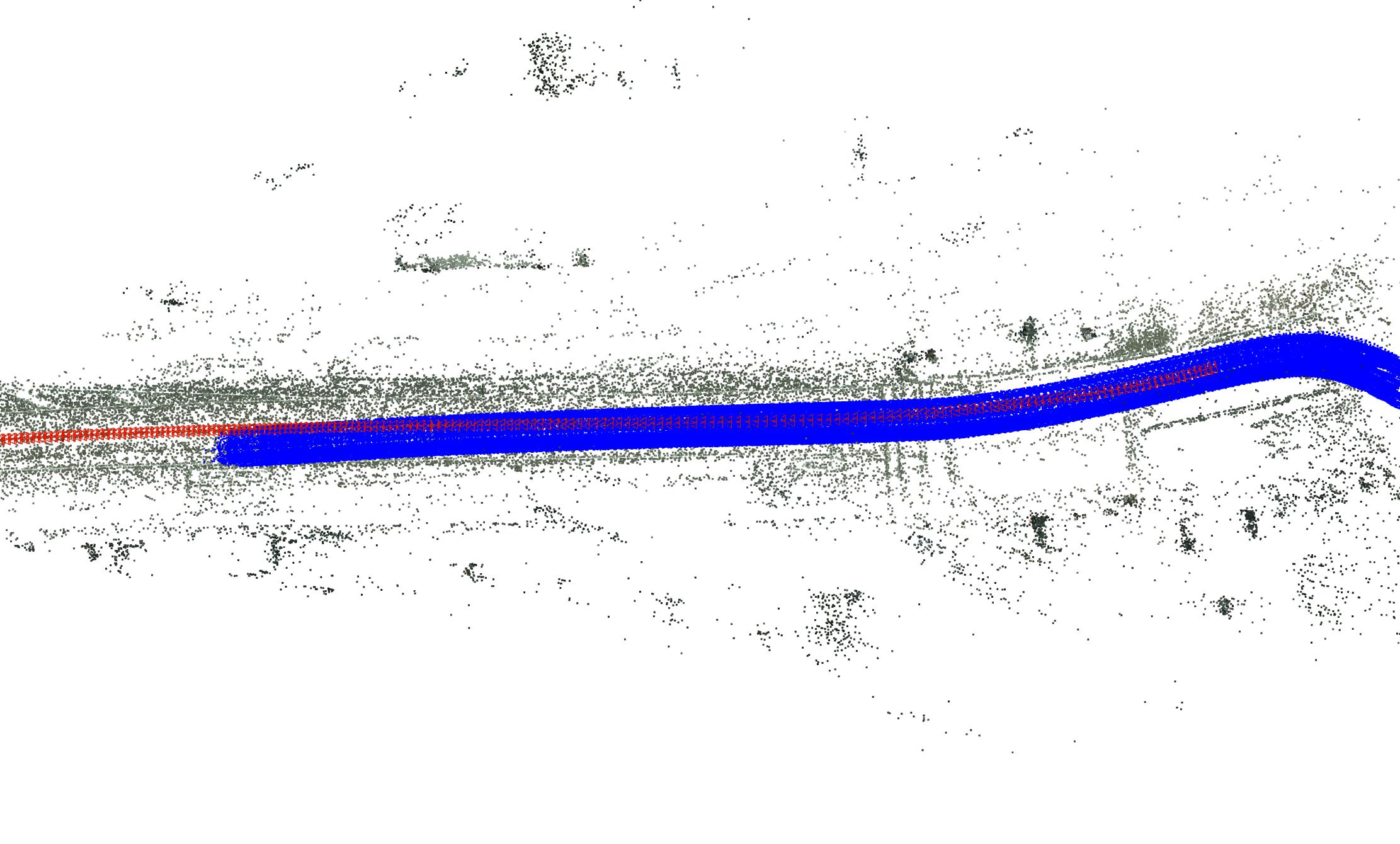}
		\hfill
		\includegraphics[width = \subfigureWidthFourColumns,frame]{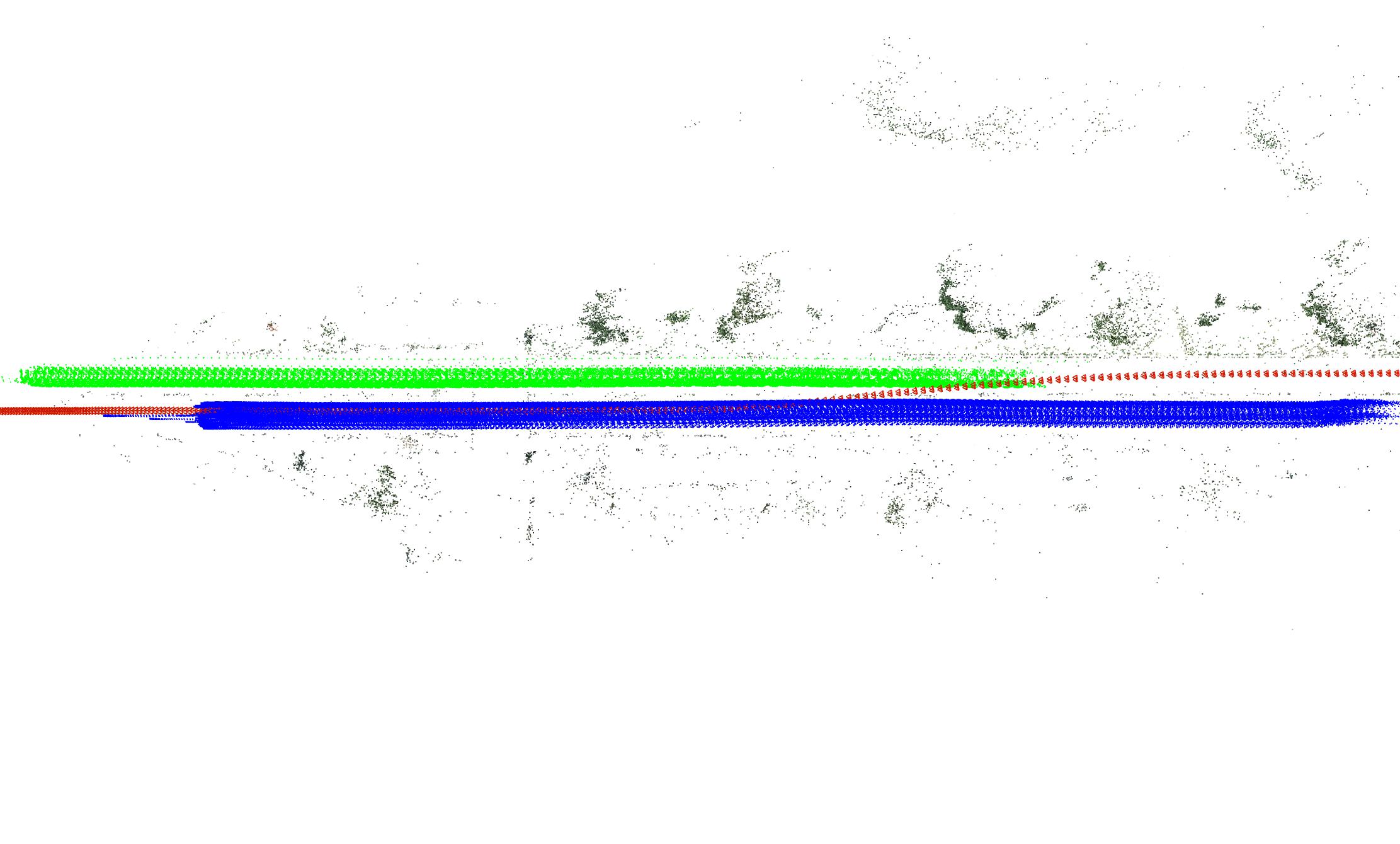}
		\hfill
		\includegraphics[width = \subfigureWidthFourColumns,frame]{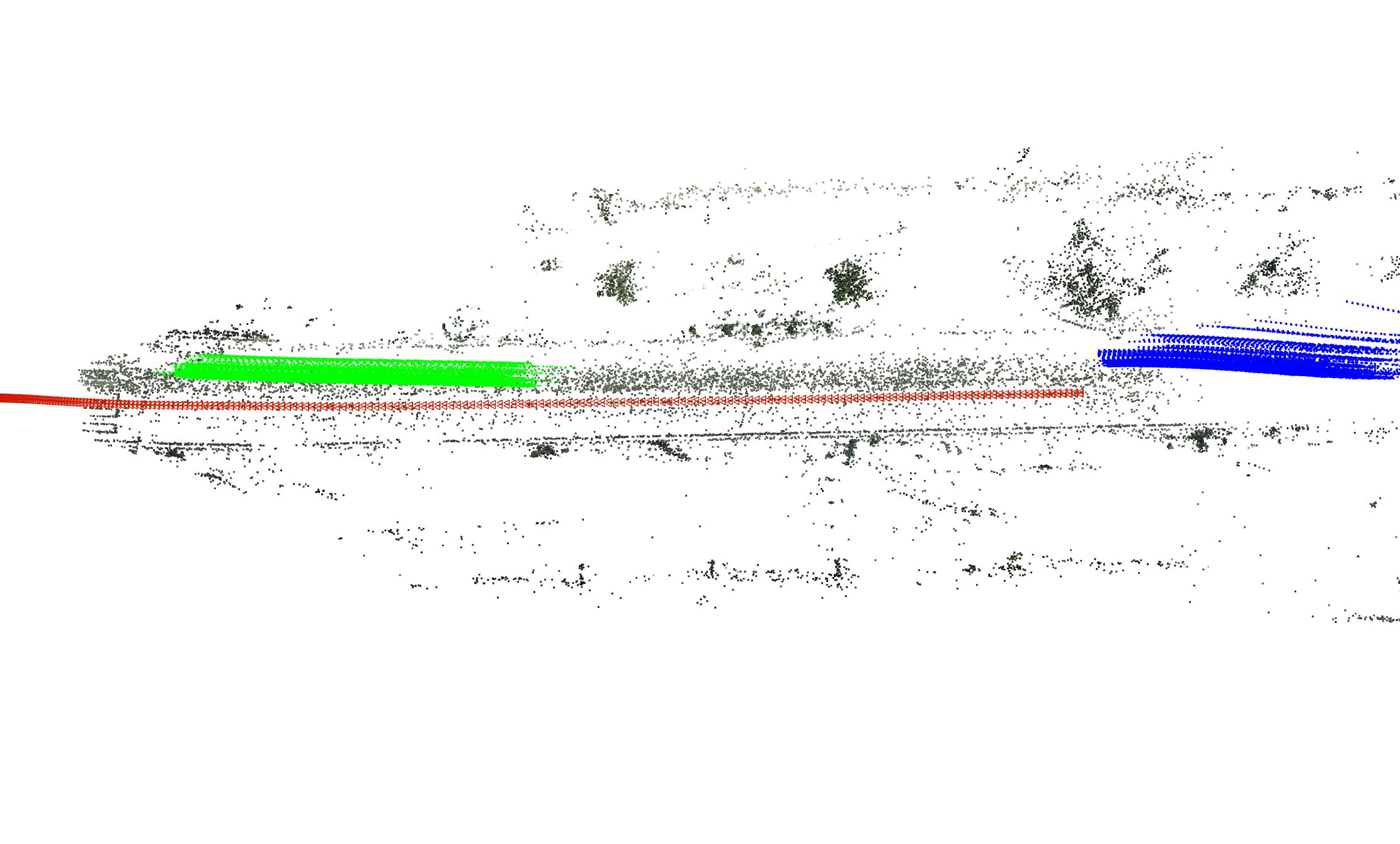}
		\hfill
		\includegraphics[width = \subfigureWidthFourColumns,frame]{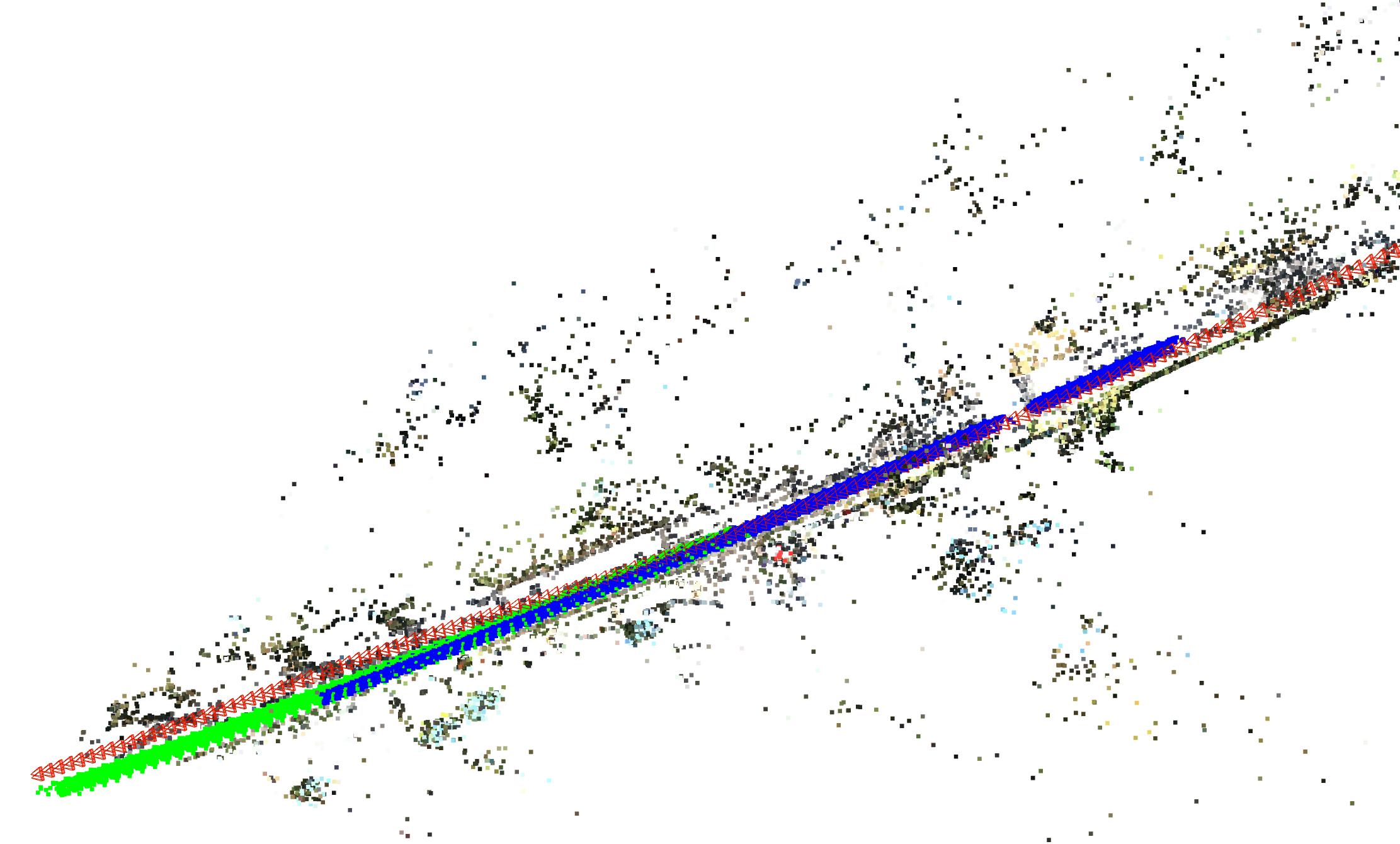}
		\caption{Trajectory Reconstruction (Top View).}
		\label{fig:qual_trajectory_top_view}
	\end{subfigure}
	\hfill
	\begin{subfigure}[t]{\textwidth}
		\includegraphics[width = \subfigureWidthFourColumns,frame]{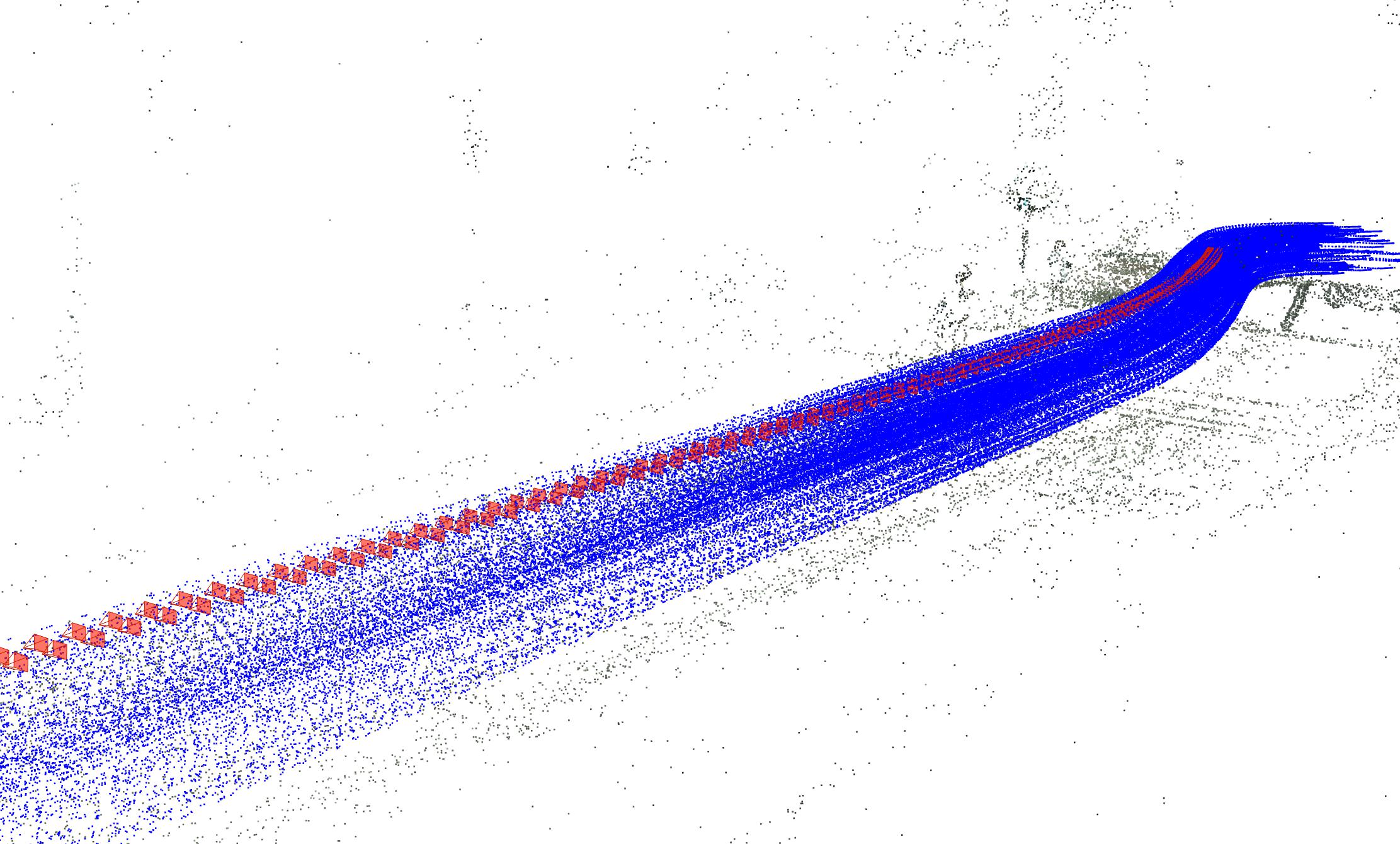}
		\hfill
		\includegraphics[width = \subfigureWidthFourColumns,frame]{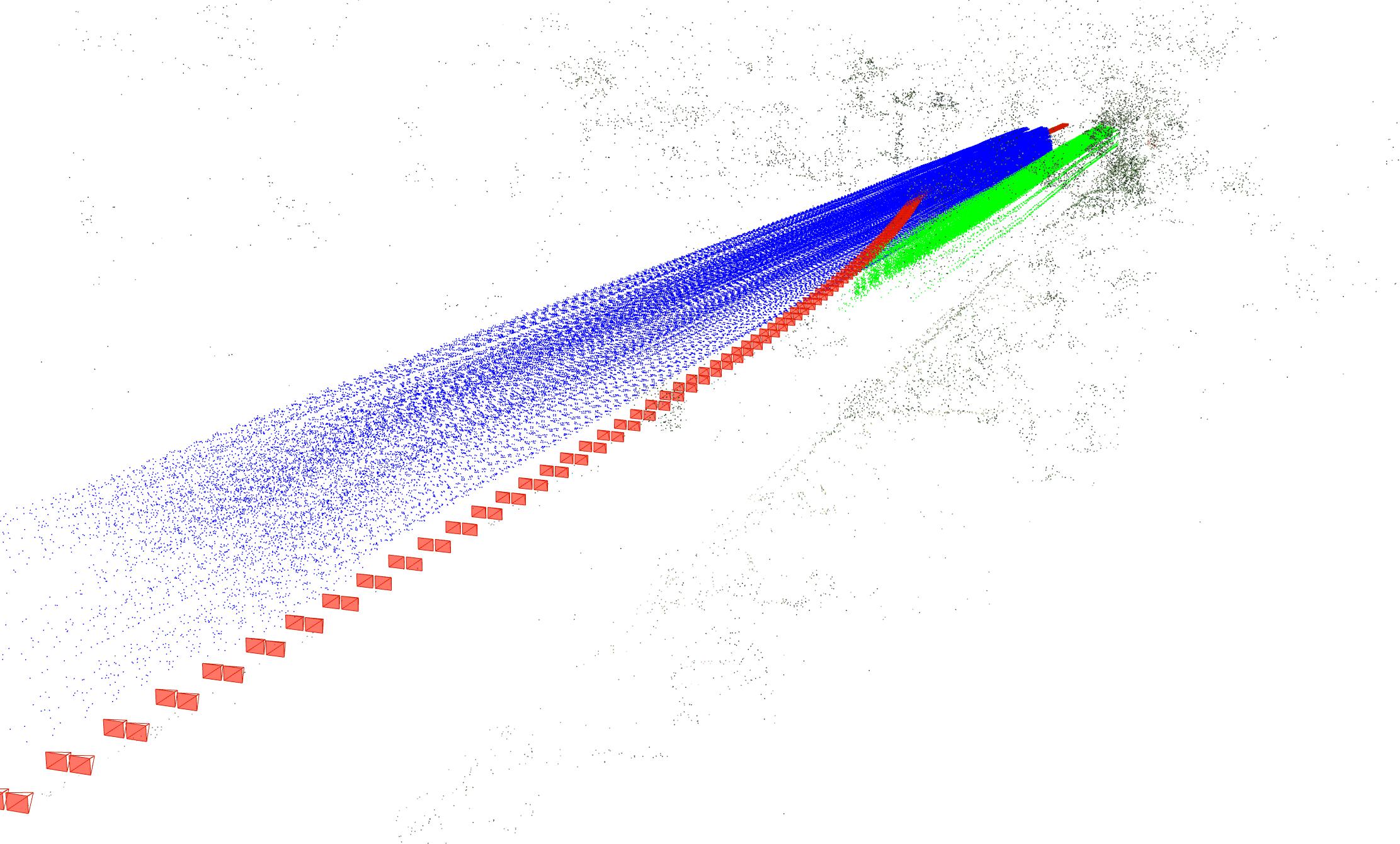}
		\hfill
		\includegraphics[width = \subfigureWidthFourColumns,frame]{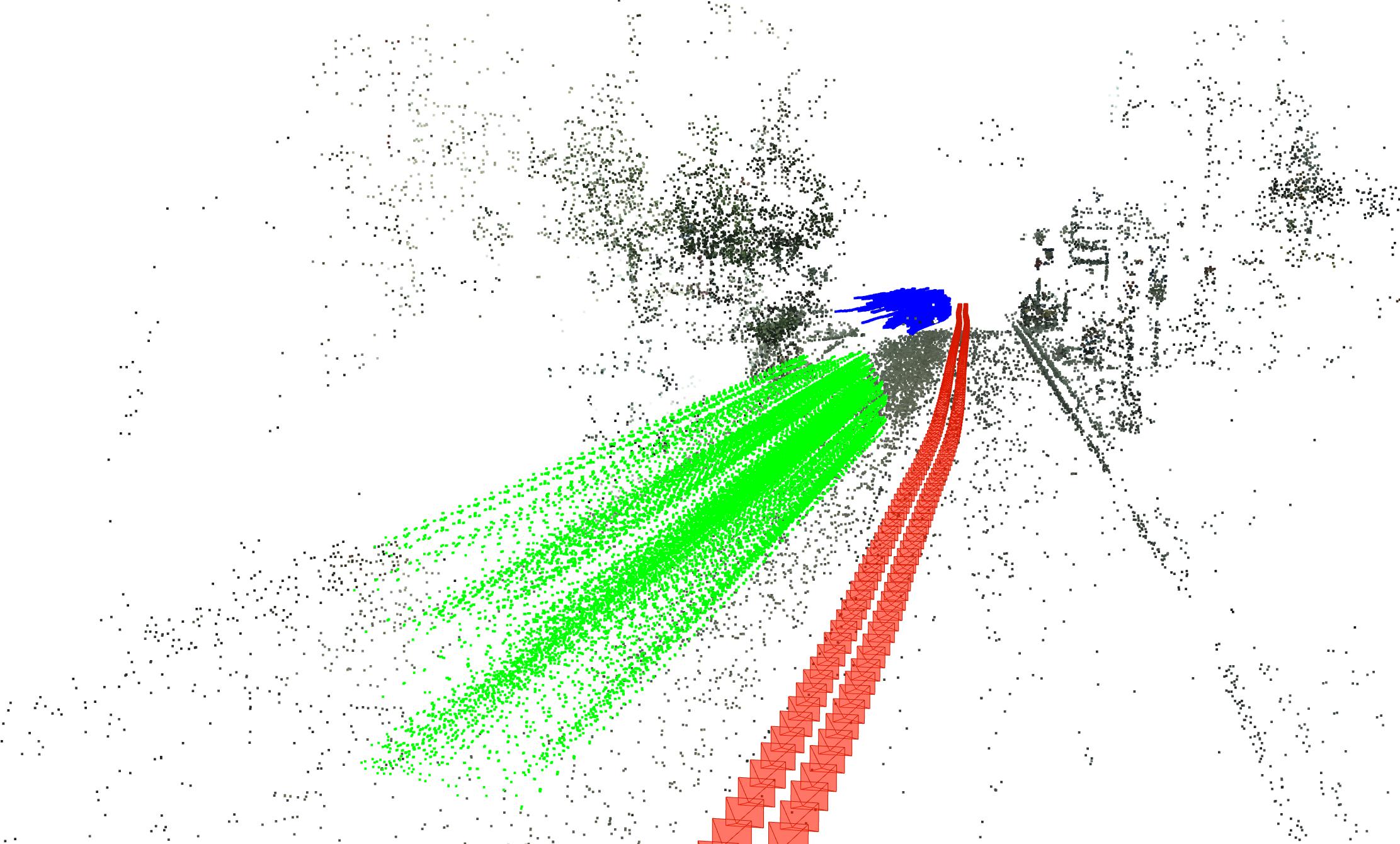}
		\hfill
		\includegraphics[width = \subfigureWidthFourColumns,frame]{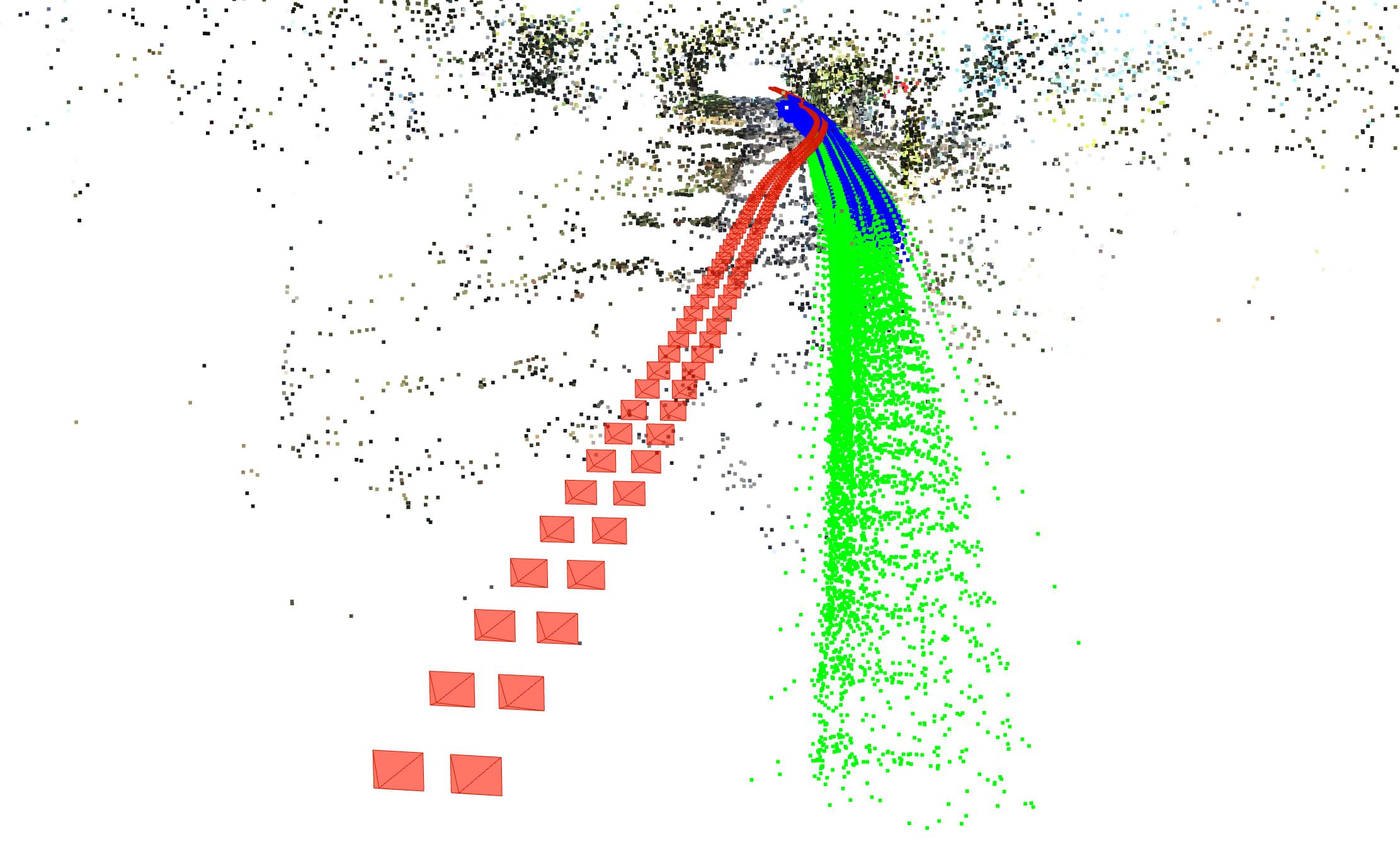}
		\caption{Trajectory Reconstruction (Side View).}
		\label{fig:qual_trajectory_side_view}
	\end{subfigure}
	\caption{Vehicle trajectory reconstruction using three sequences (stuttgart01-stuttgart03) contained in the Cityscape dataset \cite{Cordts2016Cityscapes} and one sequence (2011\_09\_26\_drive\_0013) of the KITTI dataset \cite{Geiger2013Kitti}. Object segmentations and object reconstructions are exemplarily shown for one of the vehicles visible in the scene. The reconstructed cameras are shown in red. The vehicle trajectories are colored in green and blue. The figure is best viewed in color.}
	\label{fig:evaluation:qual_evaluation}
\end{figure*}

%% file: experiments.tex
\section{Experiments and Evaluation}
\label{section:experiments_and_evaluation}

Due to the lack of suitable benchmark datasets, we show qualitative results using publicly available video data \cite{Cordts2016Cityscapes,Geiger2013Kitti}. For object tracking we evaluated \cite{Dai2016,Li2016,HeGDG17} for instance-aware semantic segmentation and \cite{Hu2016,Ilg2017} for optical flow computations. We observed that \cite{HeGDG17} and \cite{Hu2016} achieved the best segmentation and optical flow results. \cite{Hu2016} computes more stable optical flow vectors for moving objects than \cite{Ilg2017}. We considered the following SfM pipelines for object and background reconstructions: Colmap \cite{Schoenberger2016sfm}, OpenMVG \cite{Moulon2013}, Theia \cite{Sweeney2014} and VisualSfM \cite{Wu2011}. Our object trajectory reconstruction pipeline uses Colmap for object and OpenMVG for background reconstructions. Colmap and OpenMVG created the most reliable object and background reconstructions in our experiments.

%quantitative evaluations: Use known baseline distance (in meter) [OF THE BACKGROUND CAMERAS (should be stable)] to convert the point cloud triangulation error of left and right images to the scale of the real world

%% file: conclusion.tex
\section{Conclusions}	
\label{section:conclusions}
 
This paper presents a pipeline to reconstruct the three-dimensional trajectory of moving objects using stereo video data. We presented a novel approach to track objects on pixel level across stereo video sequences. This allows us to apply state-of-the-art SfM techniques simultaneously to different objects. We demonstrate how to resolve the scale ambiguity of object and background sfm reconstructions leveraging stereo constraints. In contrast to previously published stereo 3D object trajectory reconstruction approaches, our method leverages temporal adjacent frames for object and background reconstruction. Thus, the presented method is not limited by the stereo camera baseline. Due to the lack of stereo 3D object motion trajectory benchmark datasets with suitable ground truth data, we showed qualitative results on the Cityscape and the KITTI dataset. In future work we will analyze robustness and limitations of the presented approach w.r.t decreasing object sizes.

% object reconstruction must be improved

%\subsection{Limitations and Future Work}
%We observe that our current segmentation pipeline fails to segment objects correctly seen from a bird's eye perspective. This is due to the fact that these views are missing in the training data. 
%Finally, it is important to note that the quality of the derived information is relative to the camera as well as the object motion. Moreover, our method is subject to general SfM constraints. \\